\documentclass[11pt,a4paper]{article}

\usepackage[utf8]{inputenc}
\usepackage[T1]{fontenc}
\usepackage{amsmath,amssymb,amsfonts}
\usepackage{hyperref}
\usepackage[capitalize]{cleveref}
\usepackage{booktabs}
\usepackage{tikz}
\usetikzlibrary{arrows.meta,positioning,calc,fit,backgrounds}
\usepackage{listings}
\usepackage{xcolor}
\usepackage{enumitem}
\usepackage{geometry}

\usepackage{xurl}
\usepackage{csquotes}
\usepackage{float}

\newcommand{\code}[1]{%
  {\urlstyle{tt}\url{#1}}%
}

\newcommand{\yhat}{\hat{y}}
\newcommand{\yphat}{\hat{y}'}

\hypersetup{
  pdftitle={Automatic Differentiation for 
    Physics-Informed Neural Networks: From 
    Computational Graph to Parameter Gradient},
  pdfauthor={Abdeladhim Tahimi},
  colorlinks=true,
  linkcolor=blue!70!black,
  citecolor=blue!70!black,
  urlcolor=blue!70!black
}

\title{Automatic Differentiation from Scratch: How Pytorch Computes Gradients in Physics-Informed Neural Networks}
\author{Abdeladhim Tahimi\\[4pt]
  \small CECA, Universidade Federal de Alagoas (UFAL),
  Rio Largo, Alagoas, 57100-000, Brazil\\[2pt]
  \small \texttt{abdeladhim.tahimi@ceca.ufal.br}}
\date{}

\begin{document}

\maketitle

% --- Abstract ---
\begin{abstract}
\noindent
This paper traces, with explicit numerical values, 
how PyTorch's automatic differentiation (AD) engine 
computes gradients for Physics-Informed Neural 
Network (PINN) training --- a setting that requires 
two levels of differentiation: computing the physics 
derivative $\hat{y}'(t) = d\hat{y}/dt$ through the 
network, and computing parameter gradients 
$\nabla_\theta L$ of a loss that itself depends on 
$\hat{y}'(t)$. Using a 1-3-3-1 multilayer perceptron 
and the initial value problem 
$y'(t) + y(t) = 0$, $y(0) = 1$ 
from~\cite{tahimi2026physicsinformedneuralnetworksdidactic}, 
we trace the complete pipeline at every node: the 
computational graph built during the forward pass, 
the reverse-mode backward traversal that computes 
all 22~parameter gradients in a single pass, and the 
graph-on-graph mechanism by which 
\texttt{create\_graph=True} enables correct 
differentiation through the physics-informed 
residual. Every adjoint value is verified against 
the hand derivations 
of~\cite{tahimi2026physicsinformedneuralnetworksdidactic}, 
connecting the $P/Q$ sensitivity framework to the 
vector--Jacobian products used by PyTorch's autograd 
engine.
\end{abstract}



\bigskip

% ============================================================
% SECTIONS
% ============================================================

% ============================================================
% SECTION 1: INTRODUCTION
% ============================================================

\section{Introduction}
\label{sec:introduction}

% --------------------------------------------------------
% P1: Scope --- concrete entry and motivating questions
% --------------------------------------------------------

A typical Physics-Informed Neural Network (PINN) 
training loop fits in fewer than twenty lines of 
PyTorch. The forward pass computes the network output 
$\yhat(t;\theta)$; a call to 
\code{torch.autograd.grad} with 
\texttt{create\_graph=True} computes the time 
derivative $\yphat(t)$; the ODE residual 
$R = \yphat + \yhat$ is squared to form the loss; and 
\code{loss.backward()} produces all parameter 
gradients in a single call. The entire pipeline relies 
on automatic differentiation (AD) --- yet what does 
each of these calls actually compute, and why is each 
flag necessary? Answering these questions requires 
opening the black box.

% --------------------------------------------------------
% P2a: The two derivatives (problem and tasks)
% --------------------------------------------------------

\paragraph{Two differentiation tasks.}
Training a PINN requires two fundamentally different 
derivatives. Consider the first-order ODE 
$y'(t) + y(t) = 0$ with initial condition $y(0) = 1$. 
The PINN loss at a collocation point $t_c$ is
\begin{equation}\label{eq:intro_loss}
L = \underbrace{\left(
  \yhat(t_c;\theta)
  + \frac{\partial\yhat}{\partial t}\bigg|_{t_c}
\right)^2}_{L_R:\;\text{ODE residual}}
+ \;\lambda\,
\underbrace{\bigl(\yhat(0;\theta) - 1\bigr)^2}_{%
  L_{IC}:\;\text{initial condition}}.
\end{equation}
\noindent The loss involves the \emph{physics 
derivative} $\partial\yhat/\partial t$ --- how the 
network output changes when the input $t$ varies, 
with all parameters held fixed --- and training 
requires a second derivative, the \emph{training 
gradient} $\nabla_\theta L$, which describes how the 
loss changes when a parameter $\theta_k$ is adjusted. 
We write $\yphat$ for $d\yhat/dt$; since the 
parameters $\theta$ do not depend on $t$, this 
coincides with $\partial\yhat/\partial t$.

% --------------------------------------------------------
% P2b: The coupling and the create_graph consequence
% --------------------------------------------------------

\noindent In standard supervised learning, the loss 
depends only on $\yhat$, and a single backward pass 
computes $\nabla_\theta L$. Here the loss depends 
on both $\yhat$ and $\yphat$, so the training 
gradient of the residual loss is
\begin{equation}\label{eq:intro_grad}
\frac{\partial L_R}{\partial\theta_k}
= 2R\left(
  \frac{\partial\yhat}{\partial\theta_k}
  + \frac{\partial\yphat}{\partial\theta_k}
\right).
\end{equation}
\noindent The second term 
$\partial\yphat/\partial\theta_k$ is the 
PINN-specific complication: a second-order mixed 
derivative $\partial^2\yhat / 
\partial\theta_k\,\partial t$ that requires 
differentiating the physics-derivative computation 
itself with respect to the parameters. In PyTorch, 
this term exists only if the physics derivative was 
computed with \texttt{create\_graph=True}; omitting 
the flag silently drops it, producing wrong 
gradients without any error message.

% --------------------------------------------------------
% P2c: The working example
% --------------------------------------------------------

\noindent This ODE is the ``hello-world'' of 
physics-informed learning --- small enough to admit 
closed-form analysis, yet rich enough to exhibit the 
two-derivative coupling above. A companion 
paper~\cite{tahimi2026physicsinformedneuralnetworksdidactic} 
develops the analytical derivation of the resulting 
training gradient for a 1-3-3-1 MLP applied to this 
problem; the present paper traces the corresponding 
machine computation inside PyTorch's AD engine.

% --------------------------------------------------------
% P3: Background --- the AD-driven shift
% --------------------------------------------------------

\paragraph{From hand-derived gradients to framework AD.}
Training neural networks to satisfy differential 
equations predates modern deep learning. Early work by 
Dissanayake and Phan-Thien~\cite{dissanayake1994neural} 
and Lagaris et al.~\cite{lagaris1998artificial} 
derived the gradient expressions required for training 
analytically, for shallow architectures with fixed 
activation functions; the resulting formulas had to be 
re-derived whenever the architecture changed. The 
emergence of general-purpose AD inside deep-learning 
frameworks --- in particular reverse-mode AD as 
exposed by PyTorch~\cite{paszke2019pytorch} --- 
removed this constraint by computing both the physics 
derivatives (such as $\partial\yhat/\partial t$) and 
the parameter gradients $\nabla_\theta L$ generically, 
regardless of network depth or activation choice. 
This shift made deep architectures tractable and 
enabled the modern PINN framework of Raissi et 
al.~\cite{raissi2019physics}, in which boundary or 
initial conditions are enforced through additional 
loss terms, as well as concurrent neural-network PDE 
solvers such as the Deep Galerkin 
method~\cite{sirignano2018dgm} and the Deep Ritz 
method~\cite{e2018deep}. Since 2019, PINNs have been 
applied to fluid 
dynamics~\cite{raissi2020hidden}, inverse 
problems~\cite{karniadakis2021physics}, and a broad 
range of scientific computing 
tasks~\cite{cuomo2022scientific,wang2023expert}.

% --------------------------------------------------------
% P4: Gap --- what is missing and why it matters
% --------------------------------------------------------

\paragraph{The pedagogical gap.}
The theoretical foundations of AD are well 
established. Griewank and 
Walther~\cite{griewank2008evaluating} provide the 
definitive treatment of forward and reverse modes; 
Baydin et al.~\cite{baydin2018automatic} survey the 
field for a machine learning audience; standard 
textbooks~\cite{goodfellow2016deep} cover 
backpropagation, the application of reverse-mode AD 
to neural network training, following Rumelhart et 
al.~\cite{rumelhart1986learning}. On the PINN side, 
introductory treatments such as 
Katsikis et al.~\cite{katsikis2022gentle} explain the 
loss formulation and training methodology, while 
peer-reviewed didactic derivations such as 
Blechschmidt and Ernst~\cite{blechschmidt2021three} 
work through the residual symbolically by chain rule 
on a single-hidden-layer network --- stopping short 
of the AD engine itself. What is missing is the 
bridge between the analytical derivation and the 
machine computation: a single document that traces, 
at the level of individual operations and explicit 
numerical values, how the AD engine handles the 
two-derivative coupling 
of~\cref{eq:intro_loss,eq:intro_grad} on a 
multi-layer network. Specifically:
\begin{itemize}[nosep]
  \item What data structure does PyTorch build during 
    the forward pass?
  \item How does a single backward traversal compute 
    all 22 parameter gradients simultaneously?
  \item What does \texttt{create\_graph=True} actually 
    do, and why does omitting it silently produce 
    wrong gradients?
  \item How do the adjoint vectors propagated by 
    PyTorch relate to the $P/Q$ sensitivities derived 
    by hand 
    in~\cite{tahimi2026physicsinformedneuralnetworksdidactic}?
\end{itemize}
\noindent Existing references address pieces of this 
story --- AD theory without PINN specifics, PINN 
formulations without AD internals --- but none 
assembles them into a single, verifiable narrative 
on a worked example. Closing this gap prevents silent 
implementation errors, clarifies the computational 
cost of physics-informed training, and provides the 
foundation for developing techniques that improve 
convergence and accuracy.

% --------------------------------------------------------
% P5: Contribution, roadmap, and scope limits
% --------------------------------------------------------

\paragraph{What this paper covers.}
We trace the complete AD pipeline for the same 
1-3-3-1 MLP and the same IVP $y' + y = 0$, 
$y(0) = 1$ used 
in~\cite{tahimi2026physicsinformedneuralnetworksdidactic}. 
This pairing is small enough that every intermediate 
quantity --- the activations of the forward pass, the 
22 parameter gradients of the backward pass, and the 
extended graph created by \texttt{create\_graph=True} 
--- can be displayed explicitly, yet large enough 
that the two-derivative coupling of 
\cref{eq:intro_loss,eq:intro_grad} appears in full 
generality:

\begin{itemize}[nosep]
  \item \Cref{sec:tangents_adjoints} introduces 
    tangent and adjoint vectors on a minimal 
    two-input composition.
  \item \Cref{sec:forward} traces the forward pass 
    of the 1-3-3-1 MLP and catalogues the 
    computational graph.
  \item \Cref{sec:backward} traces the backward pass 
    node by node, computing all 22 parameter 
    gradients in one traversal.
  \item \Cref{sec:pinn_derivatives} explains the 
    graph-on-graph: how \texttt{create\_graph=True} 
    enables correct training gradients, including 
    the product rule and $\phi''$ terms.
  \item \Cref{sec:practical} addresses common 
    pitfalls, memory, and implementation patterns.
\end{itemize}

\noindent Every intermediate value is verified 
against the hand derivations 
of~\cite{tahimi2026physicsinformedneuralnetworksdidactic} 
and against PyTorch's output; a companion Jupyter 
notebook reproduces all calculations. The physics 
derivative $\yphat$ can also be computed by 
forward-mode AD --- during the forward pass, a 
tangent vector $\dot{a}^{(\ell)}$ is carried 
alongside each activation and updated by the local 
Jacobian at each layer, exactly the dual propagation 
in the forward-pass tables 
of~\cite{tahimi2026physicsinformedneuralnetworksdidactic} 
--- and PyTorch supports this via 
\code{torch.func.jvp}. We trace the reverse-mode 
approach because it is the PyTorch default for PINN 
training. This paper does not propose a new PINN 
methodology, benchmark training strategies, or 
compare AD systems; its contribution is explanatory.
% ============================================================
% SECTION 2: TANGENTS AND ADJOINTS
% ============================================================

\section{Tangents and Adjoints: The Two Faces of the 
Chain Rule}
\label{sec:tangents_adjoints}

Automatic differentiation (AD) is a family of algorithms that compute derivatives by decomposing a computation into elementary operations and applying the chain rule systematically. PyTorch records these operations in a data structure called the \emph{computational graph}. Before tracing derivatives through a neural network, we introduce the forward pass, forward-mode AD, backward-mode AD, and the graph-on-graph mechanism on a scalar function of two inputs --- the smallest example that contains every ingredient PINN training needs.

%--------------------------------------------------------
\subsection{A Minimal Example}
\label{subsec:minimal_example}
%--------------------------------------------------------

Let $y = f(g(x_1, x_2))$ with:
\begin{equation}\label{eq:fg_def}
g(x_1, x_2) = 2x_1 + x_2, \qquad f(u) = u^2.
\end{equation}
At $x_1 = 2$, $x_2 = 1$:
\begin{equation}\label{eq:fg_values}
u = g(2,1) = 5, \qquad y = f(5) = 25.
\end{equation}
The local derivatives are:
\begin{equation}\label{eq:fg_locals}
\frac{\partial g}{\partial x_1} = 2, \qquad
\frac{\partial g}{\partial x_2} = 1, \qquad
f'(u) = 2u = 10.
\end{equation}
By the chain rule, $\partial y/\partial x_1 = 
10 \cdot 2 = 20$ and $\partial y/\partial x_2 = 
10 \cdot 1 = 10$. We first compute $y = 25$ and trace what PyTorch records, then compute these derivatives in two ways that correspond to the two modes of automatic differentiation.

%--------------------------------------------------------
\subsection{The Forward Pass and Graph~1}
\label{subsec:forward_pass_fg}
%--------------------------------------------------------

The forward pass computes $y = 25$ by evaluating the operations in sequence: $u = 2x_1 + x_2 = 5$, then $y = u^2 = 25$. When an input has \texttt{requires\_grad=True}, PyTorch builds a computational graph that records this sequence. Each operation becomes a \emph{node} in the graph --- a record containing:
\begin{itemize}[leftmargin=1.5em, itemsep=2pt]
  \item the operation performed and the numerical result (the primal value),
  \item links to the input nodes and the Jacobian entries (\textbf{gradient function} \texttt{grad\_fn}) required to compute the derivative during the backward pass.
\end{itemize}

\begin{figure}[htbp]
\centering
\includegraphics[width=0.75\textwidth]{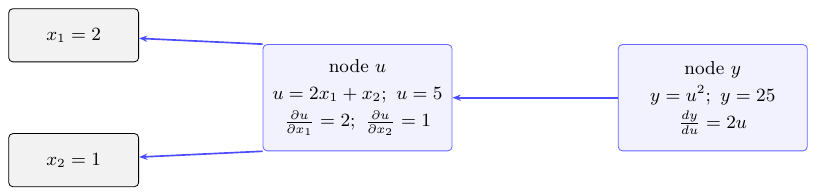}
\caption{Graph~1 built during the forward pass for $y = f(g(x_1, x_2))$. Gray boxes: leaf nodes (inputs). Blue boxes: operation nodes, each displaying the node name, the operation, the computed primal value, and the associated \emph{gradient function}. Arrows indicate dependencies, pointing from the output back to the parent nodes that fed the operation.}
\label{fig:comp_graph_fg}
\end{figure}

\Cref{fig:comp_graph_fg} shows Graph~1 built during this forward pass.

Graph~1 is the recipe for computing derivatives: it stores the operations and the numerical values needed to evaluate local derivatives during a derivative pass. No input is 
perturbed (as in finite differences), and no 
symbolic formula is manipulated --- the graph is a 
trace of elementary operations, each paired with a 
known, exact derivative rule.

\iffalse
%--------------------------------------------------------
\subsection{Forward-Mode AD: Propagating a Tangent}
\label{subsec:forward_mode}
%--------------------------------------------------------

Forward-mode AD uses Graph~1 to compute derivatives by propagating a \emph{tangent vector} $(\dot{x}_1, \dot{x}_2)$ from inputs to output --- a unit vector $(a,b) = (d{x_1}/d{\lambda}, d{x_2}/d{\lambda})$ with $\lambda$ an arbitrary parameter. At each node, the rule is 
$\dot{\text{output}} = J_{\text{local}} \cdot 
\dot{\text{input}}$ --- a \textbf{Jacobian--vector 
product (JVP)}. The result $\dot{y}$ is the 
directional derivative of $y$ in the direction of 
the seed; choosing a standard basis vector 
$e_i$ as the seed recovers the partial derivative 
$\partial y / \partial x_i$.
\fi

%--------------------------------------------------------
\subsection{Forward-Mode AD: Propagating a Tangent}
\label{subsec:forward_mode}
%--------------------------------------------------------

Forward-mode AD computes derivatives by propagating a \emph{tangent vector} $\dot{\mathbf{x}}$ from the inputs through the computational graph. By seeding the input with a vector $\mathbf{v} = (\dot{x}_1, \dot{x}_2)$, which can be viewed as $d\mathbf{x}/d\lambda$ for an independent parameter $\lambda$, the derivative is computed at each node via the rule:
\begin{equation}
    \dot{\text{output}} = J_{\text{local}} \cdot \dot{\text{input}}
\end{equation}

This operation is a \textbf{Jacobian--vector product (JVP)}, where the $m \times n$ Jacobian left-multiplies an $n \times 1$ tangent vector. This preserves the flow of data, mapping the $n$ input perturbations to $m$ output tangents. In the transition from inputs to node $u$, the $1 \times 2$ Jacobian $J_u = [\partial u/\partial x_1, \partial u/\partial x_2]$ multiplies the $2 \times 1$ tangent $(\dot{x}_1, \dot{x}_2)^\top$ to produce the scalar tangent $\dot{u}$.

The final result $\dot{y}$ represents the directional derivative of $y$ in the direction of the seed $\mathbf{v}$, scaled by its magnitude. Specifically, choosing the standard basis vector $e_i$ as the seed recovers the partial derivative $\partial y / \partial x_i$.

\Cref{tab:forward_passes} traces the two passes 
needed to obtain both partial derivatives.

\begin{table}[htbp]
\centering
\footnotesize
\caption{Forward-mode AD requires one pass per 
input. Each pass propagates a tangent seed 
through Graph~1 using the same sequence of JVPs.}
\label{tab:forward_passes}
\begin{tabular}{@{}lcc@{}}
\toprule
\textbf{Step} & 
\textbf{Pass 1:} $\dot{x}_1\!=\!1,\; 
\dot{x}_2\!=\!0$ & 
\textbf{Pass 2:} $\dot{x}_1\!=\!0,\; 
\dot{x}_2\!=\!1$ \\
\midrule
At node $u$: \;$\dot{u} = 2\dot{x}_1 + 1\cdot\dot{x}_2$ & 
  $2 \cdot 1 + 1 \cdot 0 = 2$ & 
  $2 \cdot 0 + 1 \cdot 1 = 1$ \\[3pt]
At node $y$: \;$\dot{y} = f'(u)\cdot\dot{u} = 10\,\dot{u}$ & 
  $10 \cdot 2 = 20\;\checkmark$ & 
  $10 \cdot 1 = 10\;\checkmark$ \\
\bottomrule
\end{tabular}
\end{table}

\begin{figure}[htbp]
\centering
\includegraphics[width=0.75\textwidth]{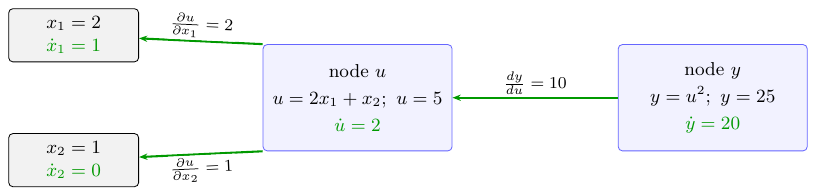}

\vspace{1em}

\includegraphics[width=0.75\textwidth]{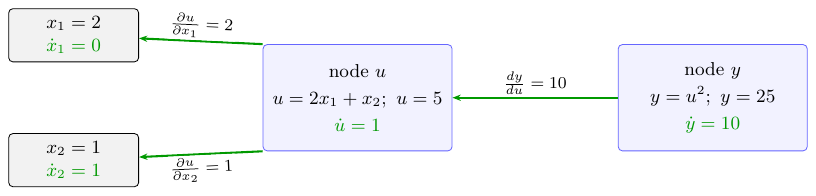}
\caption{Graph~1 traversed in forward mode. \textbf{Top:} Pass~1 with seed $\dot{x}_1 = 1, \dot{x}_2 = 0$. \textbf{Bottom:} Pass~2 with seed $\dot{x}_1 = 0, \dot{x}_2 = 1$.}
\label{fig:forward_passes_visual}
\end{figure}

\Cref{fig:forward_passes_visual} shows the two passes through Graph~1. Tangent values (green, bold) are shown inside each node. Pass~1 seeds $\dot{x}_1 = 1, \dot{x}_2 = 0$ and produces $\dot{y} = 20 = \partial y/\partial x_1$. Pass~2 seeds $\dot{x}_1 = 0, \dot{x}_2 = 1$ and produces $\dot{y} = 10 = \partial y/\partial x_2$.

%--------------------------------------------------------
\subsection{Backward-Mode AD: Propagating an Adjoint}
\label{subsec:backward_mode}
%--------------------------------------------------------

Backward-mode AD utilizes the computational graph to compute derivatives by propagating an \emph{adjoint} from the output back to the inputs. The process begins at the output with a seed $\bar{y} = dy/dy = 1$. At each node, the adjoint is propagated backward according to the rule:
\begin{equation}
    \bar{\text{input}} = \bar{\text{output}} \cdot J_{\text{local}}
\end{equation}

This is a \textbf{vector--Jacobian product (VJP)}. To propagate sensitivities backward, the $1 \times m$ adjoint left-multiplies the $m \times n$ Jacobian. This order is dimensionally required to "pull" the sensitivity of the $m$ outputs back to the $n$ inputs, resulting in a $1 \times n$ adjoint vector. Projecting this to our example, the scalar adjoint $\bar{u}$ (a $1 \times 1$ vector) multiplies the $1 \times 2$ Jacobian $J_u$ to simultaneously "distribute" the sensitivity to both $\bar{x}_1$ and $\bar{x}_2$ in a $1 \times 2$ row.

Here, the overbar notation $\bar{v} = \partial y / \partial v$ denotes the \emph{adjoint} of $v$, representing the sensitivity of the final output $y$ with respect to the intermediate variable $v$. Unlike forward-mode, a single backward pass efficiently recovers the gradient of the output with respect to all input parameters simultaneously.

\Cref{tab:backward_pass} traces the single pass.

\begin{table}[htbp]
\centering
\footnotesize
\caption{Backward-mode AD computes both gradients 
in one pass by propagating an adjoint from output 
to inputs through Graph~1.}
\label{tab:backward_pass}
\begin{tabular}{@{}llc@{}}
\toprule
\textbf{Step} & \textbf{VJP rule} & 
\textbf{Result} \\
\midrule
Seed & $\bar{y} = 1$ & $1$ \\[3pt]
At node $y$: \;$\bar{u} = \bar{y} \cdot f'(u)$ & 
  $1 \cdot 10$ & $10$ \\[3pt]
At node $u$: \;$\bar{x}_1 = \bar{u} \cdot 
\partial g/\partial x_1$ & 
  $10 \cdot 2$ & $20\;\checkmark$ \\[3pt]
At node $u$: \;$\bar{x}_2 = \bar{u} \cdot 
\partial g/\partial x_2$ & 
  $10 \cdot 1$ & $10\;\checkmark$ \\
\bottomrule
\end{tabular}
\end{table}

\noindent One pass produced both 
$\partial y/\partial x_1 = 20$ and 
$\partial y/\partial x_2 = 10$, plus the 
intermediate adjoint $\bar{u} = 10$.

\begin{figure}[htbp]
\centering
\includegraphics{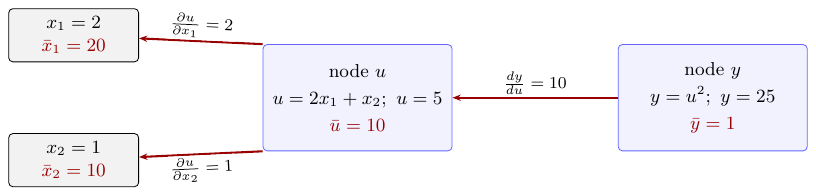}
\caption{Graph~1 traversed in backward mode. Adjoint values (red, bold) are shown inside each node.}
\label{fig:backward_pass}
\end{figure}

\Cref{fig:backward_pass} shows the same Graph~1 traversed in backward mode. The adjoint seed $\bar{y} = 1$ starts at the output node and flows right to left. At each node, the incoming adjoint is multiplied by the local derivative on each outgoing edge to produce the adjoint at the next node: $\bar{u} = \bar{y} \cdot 10 = 10$, then $\bar{x}_1 = \bar{u} \cdot 2 = 20$ and $\bar{x}_2 = \bar{u} \cdot 1 = 10$.

\paragraph{The key distinction.}
Both modes apply the same local derivatives 
($f' = 10$, $\partial g/\partial x_1 = 2$, 
$\partial g/\partial x_2 = 1$). They differ in 
direction and cost:

\begin{center}
\footnotesize
\begin{tabular}{@{}lll@{}}
\toprule
& \textbf{Forward-mode (JVP)} & 
  \textbf{Backward-mode (VJP)} \\
\midrule
Direction & input $\to$ output & 
  output $\to$ input \\
Local rule & $\dot{y} = J \cdot \dot{x}$ & 
  $\bar{x} = \bar{y} \cdot J$ \\
One pass gives & $\partial y/\partial x_i$ 
  for one $i$ & 
  $\partial y/\partial x_i$ for all $i$ \\
Cost for $n$ inputs, $m$ outputs & 
  $n$ passes & $m$ passes \\
\midrule
\textbf{Our example} ($n\!=\!2$, $m\!=\!1$) & 
  2 passes & \textbf{1 pass} \\
\bottomrule
\end{tabular}
\end{center}

For any gradient-based optimization involving a scalar loss and a high-dimensional parameter space ($m = 1, n \gg 1$), backward-mode AD is significantly more efficient as it computes the full gradient $\nabla y \in \mathbb{R}^n$ in a single backward pass. This computational advantage is the reason PyTorch adopts backward-mode as its default autodiff engine and provides the \texttt{loss.backward()} interface for gradient calculation.

%--------------------------------------------------------
\subsection{Second Order Derivative and the Graph-on-Graph Mechanism}
\label{subsec:mixed_derivative}
%--------------------------------------------------------

\begin{figure}[H]
\centering
\includegraphics{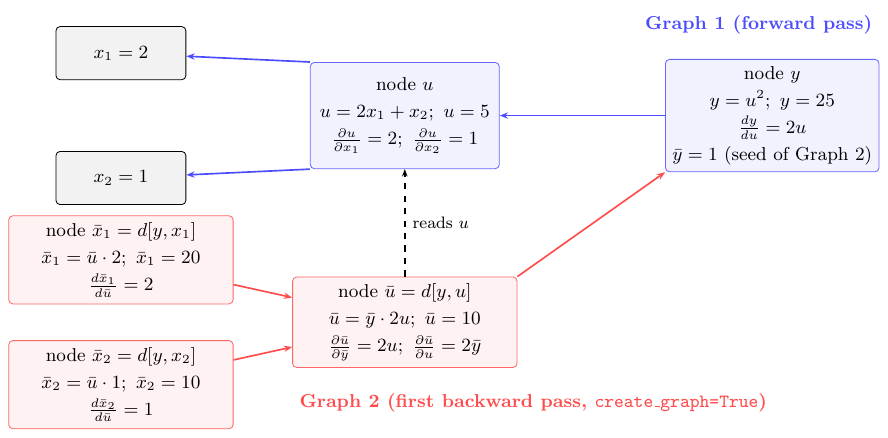}
\caption{The two graphs built during forward and backward passes. \textbf{Top:} Graph~1, built during the forward pass when \texttt{requires\_grad=True}; gray boxes are leaf nodes, blue boxes are operation nodes. \textbf{Bottom:} Graph~2, built during the first backward pass only when \texttt{create\_graph=True} is set.}
\label{fig:two_graphs}
\end{figure}

The backward pass computed $\bar{x}_1 = 20$ through VJP operations: $\bar{u} = \bar{y} \cdot 2u = 10$, then $\bar{x}_1 = \bar{u} \cdot 2 = 20$. These VJP operations are themselves arithmetic. With \texttt{create\_graph=True}, PyTorch records them in a second graph (Graph~2), making it possible to differentiate $\bar{x}_1$ with respect to the original inputs. \Cref{fig:two_graphs} shows both graphs: Graph~1 (top) built during the forward pass, and Graph~2 (bottom) built during the first backward pass. The dashed cross-link records that the VJP at $y$ reads $u = 5$ from Graph~1.

Without \texttt{create\_graph=True}, the VJP operations execute but are not recorded. The mixed derivative becomes $\partial\bar{x}_1/\partial x_2 = \partial(20)/\partial x_2 = 0$ (wrong; correct answer is 4).

\begin{figure}[H]
\centering
\includegraphics[width=0.9\textwidth]{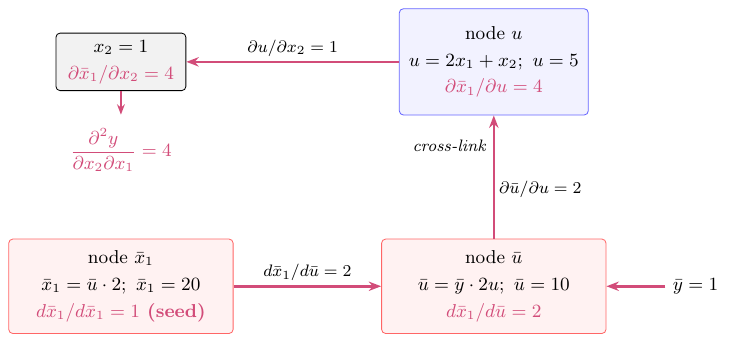}
\caption{Traversal path for computing $\partial\bar{x}_1/\partial x_2 = \dfrac{\partial^2 y}{{\partial x_2}{\partial x_1}}$.}
\label{fig:mixed_traversal}
\end{figure}

\Cref{fig:mixed_traversal} shows the traversal propagating the adjoint from $\bar{x}_1$ node with seed $1$ back to $x_2$ as a sequence of node visits, with the adjoint accumulating from 1 to 4 as it flows through the combined graph $Graph\,1 + Graph\,2$.

\paragraph{The PINN analogy.}
The roles in PINN training for the initial value problem presented in \Cref{subsec:problem} map directly onto this example:
\begin{center}
\footnotesize
\begin{tabular}{@{}ll@{}}
\toprule
\textbf{Simple example} & 
\textbf{PINN} \\
\midrule
$x_1$ & 
  time $t$ or network weight or bias $\theta$ \\
$x_2$ & 
  $\theta$ \\
$\bar{x}_1 = \partial \yhat/\partial x_1$ & 
  $\yphat = \partial \yhat/\partial t$ or ${\partial\yhat}/{\partial\theta}$\\
${\partial\bar{x}_1}/{\partial x_2}$ 
  (second backward) & 
  ${\partial\yphat}/{\partial\theta}$ 
  (needed for the residual loss) \\
Cross-link (VJP at $y$ uses $u$) & 
  Cross-links (VJP nodes use $W^{(\ell)}$, 
  $a^{(\ell)}$) \\
\bottomrule
\end{tabular}
\end{center}

%--------------------------------------------------------
\subsection{Verification in PyTorch}
\label{subsec:verify_simple}
%--------------------------------------------------------

To verify the theoretical framework, the following listings demonstrate PyTorch's tiered approach to automatic differentiation, progressing from a standard forward pass to the computation of second-order mixed derivatives. While Level 0 ignores gradient tracking, Level 1 allows for first-order gradients for the leaf tensors having \texttt{requires\_grad=True}. The use of \texttt{retain\_graph=True} in the first backward AD traversal is essential because PyTorch’s default behavior is to free the intermediate buffers once a backward pass is completed; without this flag, the computation graph would be destroyed after calculating $\bar{x}_1$, making the subsequent calculation of $\bar{x}_2$ impossible. Level 2 utilizes \texttt{create\_graph=True} to build a graph capable of tracking the second order derivatives --- the derivatives of $\bar{x}_1 = \partial y/\partial x_1$. 

\begin{lstlisting}[caption={Level~0: no 
\texttt{requires\_grad}, no derivatives.}]
import torch

x1 = torch.tensor(2.0)
x2 = torch.tensor(1.0)

u = 2 * x1 + x2
y = u ** 2

print(y.item())               #>>> 25.0
print(y.grad_fn)              #>>> None
\end{lstlisting}

\begin{lstlisting}[caption={Level~1: first 
derivatives work; mixed derivative fails.}]
x1 = torch.tensor(2.0, requires_grad=True)
x2 = torch.tensor(1.0, requires_grad=True)

u = 2 * x1 + x2
y = u ** 2

# Backward AD: x1 <- y and x2 <- y (the partial derivatives)
bar_x1 = torch.autograd.grad(y, x1,
                             retain_graph=True)[0]
bar_x2 = torch.autograd.grad(y, x2)[0]

print(bar_x1.item())          #>>> 20.0
print(bar_x2.item())          #>>> 10.0
print(bar_x1.grad_fn)         #>>> None

# Let's try the traversal x2 <- bar_x1 (the mixed derivative)
try:
    torch.autograd.grad(bar_x1, x2)
except RuntimeError as e:
    print(e)
#>>> element 0 of tensors does not require grad
#>>> and does not have a grad_fn
\end{lstlisting}

\begin{lstlisting}[caption={Level~2: 
\texttt{create\_graph=True} enables the mixed 
derivative.}]
x1 = torch.tensor(2.0, requires_grad=True)
x2 = torch.tensor(1.0, requires_grad=True)

u = 2 * x1 + x2
y = u ** 2

# traversal x1 <- y
bar_x1 = torch.autograd.grad(y, x1,
                             create_graph=True)[0]

print(bar_x1.item())                   #>>> 20.0
print(type(bar_x1.grad_fn).__name__)   #>>> MulBackward0

mixed = torch.autograd.grad(bar_x1, x2)[0]
print(mixed.item())                    #>>> 4.0
\end{lstlisting}

With the mechanics of higher-order AD established, we are ready to apply these concepts to the development of a Physics-Informed Neural Network (PINN) model designed to solve a fundamental initial value problem.
% ============================================================
% SECTION 3: THE COMPUTATIONAL GRAPH
% ============================================================

\section{The Forward Pass}
\label{sec:forward}

We now move from the two-input composition to a real 
neural network. This section defines the problem and 
the network that will serve as our running example 
for the rest of the paper, computes the forward pass 
with explicit numerical values, and traces the 
computational graph that PyTorch builds along the way.

%--------------------------------------------------------
\subsection{The Problem}
\label{subsec:problem}
%--------------------------------------------------------

We seek to approximate the solution of the initial 
value problem
\begin{equation}\label{eq:ode}
y'(t) + y(t) = 0, \qquad y(0) = 1,
\end{equation}
whose analytical solution is $y(t) = e^{-t}$. A 
Physics-Informed Neural Network (PINN) approximates 
$y(t)$ by a neural network $\yhat(t;\theta)$ and 
trains it by minimizing a loss function that penalizes 
the ODE residual $R(t) = \yphat(t) + \yhat(t)$ at 
collocation points and the initial condition error 
$\yhat(0) - 1$. The details of the loss will be 
needed in \Cref{sec:pinn_derivatives}; for now, the 
key point is that the loss depends on both $\yhat$ 
and its derivative $\yphat = d\yhat/dt$, and the AD 
engine must handle both.

%--------------------------------------------------------
\subsection{The Network}
\label{subsec:network}
%--------------------------------------------------------

The network is a \textbf{1-3-3-1 multilayer 
perceptron}: one input ($t$), two hidden layers with 
3~neurons each using $\tanh$ activation, and one 
linear output neuron producing $\yhat(t)$. The 
forward pass through layer~$\ell$ is:
\begin{equation}\label{eq:layer_forward}
z^{(\ell)} = W^{(\ell)} a^{(\ell-1)} + b^{(\ell)}, 
\qquad
a^{(\ell)} = \phi(z^{(\ell)}),
\end{equation}
where $\phi = \tanh$ for hidden layers and 
$\phi = \text{identity}$ for the output layer. The 
network has 22~trainable parameters: 
$\theta = \{W^{(\ell)}, b^{(\ell)}\}_{\ell=1}^{3}$.

We fix all parameters to enable hand verification. 
\Cref{tab:all_params} lists the complete parameter 
set.

\begin{table}[htbp]
\centering
\footnotesize
\caption{Fixed parameters for the 1-3-3-1 MLP 
(22~parameters total).}
\label{tab:all_params}
\begin{tabular}{@{}lll@{}}
\toprule
\textbf{Layer} & \textbf{Weights} & 
\textbf{Biases} \\
\midrule
$\ell = 1$ \; ($1 \to 3$) &
$W^{(1)} = \begin{pmatrix} 
0.2 \\ -0.5 \\ 0.8 \end{pmatrix}$ &
$b^{(1)} = \begin{pmatrix} 
-0.1 \\ 0.3 \\ -0.2 \end{pmatrix}$ \\[14pt]
$\ell = 2$ \; ($3 \to 3$) &
$W^{(2)} = \begin{pmatrix} 
0.1 & -0.3 & 0.5 \\ 
0.6 & 0.2 & -0.4 \\ 
-0.2 & 0.7 & 0.1 \end{pmatrix}$ &
$b^{(2)} = \begin{pmatrix} 
0.2 \\ -0.1 \\ 0.4 \end{pmatrix}$ \\[14pt]
$\ell = 3$ \; ($3 \to 1$) &
$W^{(3)} = \begin{pmatrix} 
0.9 & -0.6 & 0.3 \end{pmatrix}$ &
$b^{(3)} = \begin{pmatrix} 
-0.3 \end{pmatrix}$ \\
\bottomrule
\end{tabular}
\end{table}

%--------------------------------------------------------
\subsection{Forward Pass at \texorpdfstring{$t = 0.5$}{y-hat prime}}
\label{subsec:forward_pass}
%--------------------------------------------------------

We trace the forward pass at the collocation point 
$t_c = 0.5$. The input is $a^{(0)} = 0.5$.

\paragraph{Layer 1.}
\begin{equation}\label{eq:z1}
z^{(1)} = W^{(1)} (0.5) + b^{(1)}
= \begin{pmatrix} 
0.1 - 0.1 \\ -0.25 + 0.3 \\ 0.4 - 0.2 
\end{pmatrix}
= \begin{pmatrix} 0.0 \\ 0.05 \\ 0.2 \end{pmatrix},
\qquad
a^{(1)} = \tanh(z^{(1)})
= \begin{pmatrix} 
0.0000 \\ 0.0500 \\ 0.1974 \end{pmatrix}.
\end{equation}

\paragraph{Layer 2.}
\begin{equation}\label{eq:z2}
z^{(2)} = W^{(2)} a^{(1)} + b^{(2)}
= \begin{pmatrix} 
0.2837 \\ -0.1690 \\ 0.4547 \end{pmatrix},
\qquad
a^{(2)} = \tanh(z^{(2)})
= \begin{pmatrix} 
0.2763 \\ -0.1673 \\ 0.4257 \end{pmatrix}.
\end{equation}
To verify one entry: $z^{(2)}_1 = 0.1(0.0000) + 
(-0.3)(0.0500) + 0.5(0.1974) + 0.2 
= 0 - 0.0150 + 0.0987 + 0.2 = 0.2837$.

\paragraph{Layer 3 (output).}
\begin{equation}\label{eq:z3}
\yhat = W^{(3)} a^{(2)} + b^{(3)}
= 0.9(0.2763) + (-0.6)(-0.1673) + 0.3(0.4257) - 0.3
= 0.1768.
\end{equation}

\noindent \Cref{tab:forward_values} collects the 
complete forward pass.

\begin{table}[htbp]
\centering
\footnotesize
\caption{Forward pass for $t = 0.5$.}
\label{tab:forward_values}
\begin{tabular}{@{}lll@{}}
\toprule
\textbf{Layer} & \textbf{Operation} & 
\textbf{Result} \\
\midrule
Input & $a^{(0)} = t$ & $0.5$ \\[2pt]
L1 affine & $z^{(1)} = W^{(1)} a^{(0)} + b^{(1)}$ & 
  $(0.0,\; 0.05,\; 0.2)$ \\[2pt]
L1 activation & $a^{(1)} = \tanh(z^{(1)})$ & 
  $(0.0000,\; 0.0500,\; 0.1974)$ \\[2pt]
L2 affine & $z^{(2)} = W^{(2)} a^{(1)} + b^{(2)}$ & 
  $(0.2837,\; -0.1690,\; 0.4547)$ \\[2pt]
L2 activation & $a^{(2)} = \tanh(z^{(2)})$ & 
  $(0.2763,\; -0.1673,\; 0.4257)$ \\[2pt]
L3 affine & $\yhat = W^{(3)} a^{(2)} + b^{(3)}$ & 
  $0.1768$ \\
\bottomrule
\end{tabular}
\end{table}

%--------------------------------------------------------
\subsection{What PyTorch Records}
\label{subsec:node_types}
%--------------------------------------------------------

Two points deserve emphasis. First, the graph 
records \emph{elementary operations}, not neural 
network layers. A single \code{nn.Linear} layer 
produces at least two operation nodes: one for the 
matrix multiply, one for the bias addition. Second, 
the leaf nodes now include both the input $t$ and 
the parameter tensors $W^{(\ell)}$, $b^{(\ell)}$. 
In PyTorch, parameters are leaves with 
\code{requires_grad=True} by default; the input 
$t$ becomes a tracked leaf only if the user sets 
\code{requires_grad=True} explicitly --- which 
is necessary for computing $\yphat$.

\Cref{tab:graph_nodes} lists every operation node 
that PyTorch creates during the forward pass of 
\Cref{tab:forward_values}. For each node, the table 
shows the tensor it produces, the \code{grad_fn} 
that PyTorch attaches to it, and the tensors it 
saves for the backward pass.

\begin{table}[htbp]
\centering
\footnotesize
\caption{Forward pass trace (Graph 1 construction) for $t = 0.5 \mapsto \yhat = 0.1768$. Each primal node materializes an intermediate value and buffers the local Jacobian factors required for the first-order backward pass.}
\label{tab:graph_nodes}
\begin{tabular}{@{}llll@{}}
\toprule
\textbf{Primal Node} & \textbf{Produces (Operation = Value)} & \textbf{\code{grad\_fn}} & \textbf{Saved for Backward} \\
\midrule
L1 Matmul & $m^{(1)} = t (W^{(1)})^\top = (0.1, -0.25, 0.4) \in \mathbb{R}^{1 \times 3}$ & \code{MmBackward0} & $t, W^{(1)}$ \\[2pt]
L1 Add    & $z^{(1)} = m^{(1)} + b^{(1)} = (0.0, -0.05, 0.2) \in \mathbb{R}^{1 \times 3}$ & \code{AddBackward0} & (none) \\[2pt]
L1 Tanh   & $a^{(1)} = \tanh(z^{(1)}) = (0.0, -0.05, 0.19) \in \mathbb{R}^{1 \times 3}$ & \code{TanhBackward0} & $a^{(1)}$ (for $\phi'$) \\
\midrule
L2 Matmul & $m^{(2)} = a^{(1)} (W^{(2)})^\top = (0.1, 0.0, -0.01) \in \mathbb{R}^{1 \times 3}$ & \code{MmBackward0} & $a^{(1)}, W^{(2)}$ \\[2pt]
L2 Add    & $z^{(2)} = m^{(2)} + b^{(2)} = (1.1, -0.8, 0.4) \in \mathbb{R}^{1 \times 3}$ & \code{AddBackward0} & (none) \\[2pt]
L2 Tanh   & $a^{(2)} = \tanh(z^{(2)}) = (0.8, -0.66, 0.38) \in \mathbb{R}^{1 \times 3}$ & \code{TanhBackward0} & $a^{(2)}$ (for $\phi'$) \\
\midrule
L3 Matmul & $m^{(3)} = a^{(2)} (W^{(3)})^\top = 0.8768 \in \mathbb{R}$ & \code{MmBackward0} & $a^{(2)}, W^{(3)}$ \\[2pt]
L3 Add    & $\yhat = m^{(3)} + b^{(3)} = 0.1768 \in \mathbb{R}$ & \code{AddBackward0} & (none) \\
\bottomrule
\end{tabular}
\end{table}

\paragraph{Three observations.}
\begin{enumerate}
    \item \textbf{The saved tensors are the chain rule factors.} Consider the node \textit{L3 matmul}, which computes $z^{(3)} = a^{(2)}W^{(3)}$: it saves the input activation $a^{(2)}$ and the weight matrix $W^{(3)}$. In the backward pass, these tensors serve as the local Jacobian factors for two distinct VJPs. The incoming adjoint $\bar{z}^{(3)}$ is multiplied by $a^{(2)}$ to produce the weight gradient $\bar{W}^{(3)} = \partial L / \partial W^{(3)}$ for the optimizer, and multiplied by $W^{(3)}$ to propagate the sensitivity back as the activation adjoint of the previous layer $\bar{a}^{(2)} = \partial L / \partial a^{(2)}$.

    \item \textbf{Tanh saves its output, not its input.} The VJP of $\tanh$ requires the derivative $\phi'(z) = 1 - \tanh^2(z) = 1 - a^2$, which can be reconstructed from the output~$a$ alone. For example, at node \textit{L2 tanh}: $\phi'(z^{(2)}) = 1 - (a^{(2)})^2 = (0.9237,\; 0.9720,\; 0.8188)$. This is a deliberate design choice: saving $a$ instead of $z$ avoids recomputing the expensive exponential functions within $\tanh$ during the backward pass.

    \item \textbf{Addition nodes record nothing.} Nodes like \textit{L3 add} record nothing \texttt{(none)}. Since the partial derivative of an addition $z = u + b$ is simply $1$ with respect to both inputs (where $u$ represents the product term and $b$ the bias), the incoming adjoint $\bar{z}$ is passed back unchanged: $\bar{u} = \bar{z} \cdot 1$ to continue the backward flow, and $\bar{b} = \bar{z} \cdot 1$ to serve as the gradient for the optimizer. No records from the forward pass are needed to compute these adjoints and no actual calculation is required, which minimizes both memory overhead and computational effort.
\end{enumerate}

%--------------------------------------------------------
\subsection{The Complete DAG}
\label{subsec:dag}
%--------------------------------------------------------

\begin{figure}[htbp]
\centering
\includegraphics[width=0.95\textwidth]{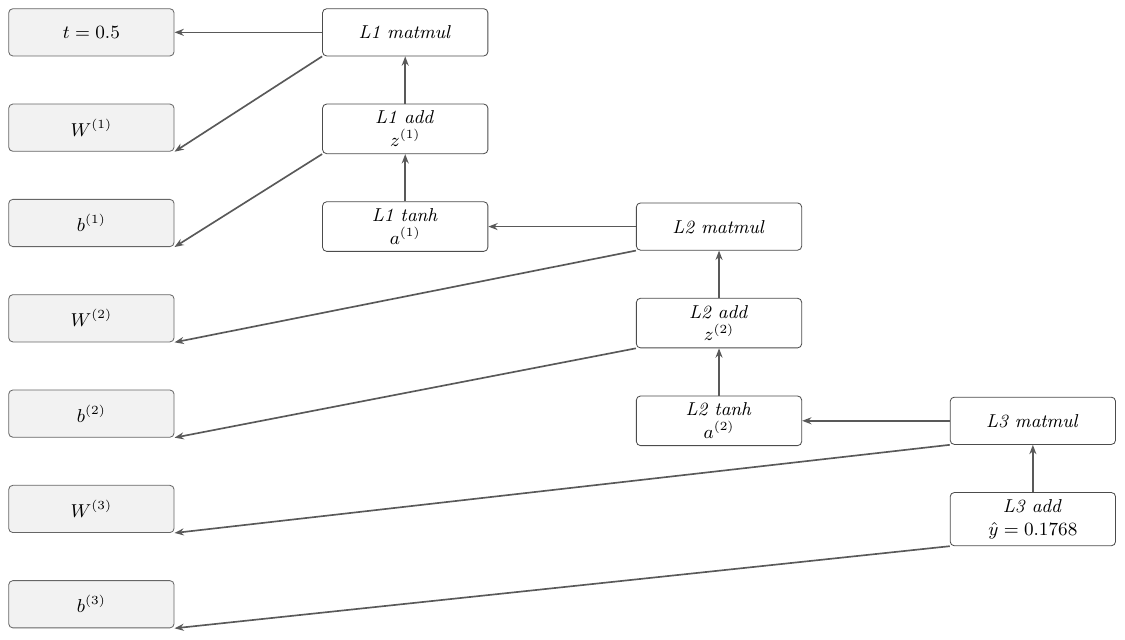}
\caption{Computational graph for 
$t = 0.5 \mapsto \yhat = 0.1768$. Gray boxes: leaf 
nodes. White boxes: operation nodes. Arrows point in the direction of adjoint flow.}
\label{fig:comp_graph}
\end{figure}

\Cref{fig:comp_graph} assembles all the nodes from 
\Cref{tab:graph_nodes} into the full directed 
acyclic graph. The gray boxes on the left are the 
leaf nodes (input and parameters); the white boxes 
are the operation nodes. Arrows indicate dependencies --- the direction the adjoint will flow when we traverse this graph backward in \Cref{sec:backward}.

%--------------------------------------------------------
\subsection{Inspecting the Graph in PyTorch}
\label{subsec:inspect}
%--------------------------------------------------------

The computational graph is not just a conceptual 
diagram --- it is a data structure that PyTorch 
builds in memory and that the user can inspect 
directly. The following listings provide a step-by-step walkthrough of the graph's instantiation and the programmatic inspection of its internal operation nodes.

\paragraph{Defining the network.}
We construct the neural network model and set the fixed 
parameters from \Cref{tab:all_params}:

\begin{lstlisting}[caption={Network definition and 
parameter initialization.}]
import torch
import torch.nn as nn

net = nn.Sequential(
    nn.Linear(1, 3), nn.Tanh(),
    nn.Linear(3, 3), nn.Tanh(),
    nn.Linear(3, 1),
)

with torch.no_grad():
    net[0].weight.copy_(torch.tensor(
        [[ 0.2],[-0.5],[ 0.8]]))
    net[0].bias.copy_(torch.tensor(
        [-0.1, 0.3,-0.2]))
    net[2].weight.copy_(torch.tensor(
        [[ 0.1,-0.3, 0.5],
         [ 0.6, 0.2,-0.4],
         [-0.2, 0.7, 0.1]]))
    net[2].bias.copy_(torch.tensor(
        [ 0.2,-0.1, 0.4]))
    net[4].weight.copy_(torch.tensor(
        [[ 0.9,-0.6, 0.3]]))
    net[4].bias.copy_(torch.tensor([-0.3]))
\end{lstlisting}

\paragraph{Forward pass and graph inspection.}
Running the forward pass builds the graph. We then inspect the output node and its parents:

\begin{lstlisting}[caption={Forward pass and node inspection.}]
t = torch.tensor([[0.5]], requires_grad=True)
y_hat = net(t)

print(y_hat.item())
#>>> 0.1768

print(type(y_hat.grad_fn).__name__)
#>>> AddBackward0

print([(type(p[0]).__name__, p[1])
       for p in y_hat.grad_fn.next_functions])
#>>> [('MmBackward0', 0), ('AccumulateGrad', 0)]

print(t.grad_fn, t.requires_grad)
#>>> None True
\end{lstlisting}

The output node \texttt{AddBackward0} represents the final addition in the network. Its \texttt{next\_\allowbreak functions} attribute contains a list of its parent nodes: the \textit{L3 matmul} and the bias leaf. These are stored as tuples \texttt{(function, index)}, where the index (0) specifies which output of the parent node feeds the current operation—a detail required for operations that return multiple tensors, which is not the case here. The loop iterates through this list to propagate gradients back through all input branches. While the bias appears as \texttt{AccumulateGrad} to manage its update, the input $t$ returns \texttt{None} for its \texttt{grad\_fn} because it is a user-defined root where the graph terminates.

\paragraph{Walking the full graph.}
Following the first parent at each node traces the 
path from the output back to the input:

\begin{lstlisting}[caption={Walking the 
computational graph from output to input.}]
node = y_hat.grad_fn
while node is not None:
    print(type(node).__name__)
    parents = node.next_functions
    node = parents[0][0] if parents else None
#>>> AddBackward0       <- L3 add
#>>> MmBackward0        <- L3 matmul
#>>> TanhBackward0      <- L2 tanh
#>>> AddBackward0       <- L2 add
#>>> MmBackward0        <- L2 matmul
#>>> TanhBackward0      <- L1 tanh
#>>> AddBackward0       <- L1 add
#>>> MmBackward0        <- L1 matmul
#>>> AccumulateGrad     <- leaf (t)
\end{lstlisting}

\noindent The sequence matches 
\Cref{tab:graph_nodes} exactly: eight operation 
nodes, terminating at the leaf. This is the recipe 
that \texttt{loss.backward()} will follow in 
\Cref{sec:backward} --- the graph stores the 
sequence of VJP operations, not the gradients 
themselves.          % label: sec:forward
% ============================================================
% SECTION 4: REVERSE-MODE AD
% ============================================================

\section{Reverse-Mode AD}
\label{sec:backward}

We now traverse the graph from \Cref{fig:comp_graph} 
backward, propagating adjoint vectors from the output 
to the leaves. By the end of this single traversal, 
we will have computed $\partial\yhat/\partial\theta_k$ 
for all 22~parameters simultaneously.

%--------------------------------------------------------
\subsection{Starting Point: The Adjoint Seed}
\label{subsec:seed}
%--------------------------------------------------------

To illustrate reverse-mode in isolation, we first 
differentiate $\yhat$ itself (not the loss). The 
adjoint seed is:
\begin{equation}\label{eq:adjoint_seed}
\bar{\yhat} = \frac{\partial\yhat}{\partial\yhat} = 1.
\end{equation}
This single number enters the graph at the output 
node and flows backward. At each operation node, the 
VJP rule uses the tensors saved during the forward 
pass (\Cref{tab:graph_nodes}) to propagate the 
adjoint one step further.

%--------------------------------------------------------
\subsection{Constructing VJP Nodes During the Backward Pass}
\label{subsec:backward_trace}
%--------------------------------------------------------
Following the conceptual framework of Graph 2 established in \Cref{subsec:mixed_derivative}, we now trace the backward pass node-by-node, formalizing the record of each Vector-Jacobian Product (VJP) as an explicit node.

\subsubsection{VJP L3 Add}

In the primal forward pass, the output layer computes $\yhat = m^{(3)} + b^{(3)}$. The node \textit{VJP L3 Add} receives the incoming adjoint $\bar{\yhat} = 1 \in \mathbb{R}$, which serves as the seed for the backward trace. Since the output activation is the identity, $\bar{z}^{(3)} = \bar{\yhat}$. 

The edges connecting the output node to its primal parents ($m^{(3)}$ and $b^{(3)}$) represent the local Jacobians $\partial\yhat/\partial m^{(3)} = 1$ and $\partial\yhat/\partial b^{(3)} = 1$. Applying the VJP operation—multiplying the incoming adjoint by these local gradients—yields:
\begin{equation}\label{eq:vjp_add3}
\bar{m}^{(3)} = \bar{\yhat} \cdot 1 = 1, \qquad
\bar{b}^{(3)} = \bar{\yhat} \cdot 1 = 1.
\end{equation}

\noindent The resulting adjoint $\bar{b}^{(3)} \in \mathbb{R}$ is the terminal gradient for the output bias, while $\bar{m}^{(3)} \in \mathbb{R}$ is propagated backward to serve as the incoming adjoint for the \textit{VJP L3 Mm} node.

\subsubsection{VJP L3 Mm}

In the primal pass, the linear transformation is $m^{(3)} = a^{(2)} (W^{(3)})^T$. The node \textit{VJP L3 Mm} receives the incoming adjoint $\bar{m}^{(3)} = 1$. To propagate the adjoint, the node evaluates the VJPs with respect to its primal parents, $a^{(2)}$ and $W^{(3)}$:
\begin{align}
\bar{a}^{(2)} &= \bar{m}^{(3)} W^{(3)} \label{eq:vjp_mm3_a_edge} \\
&= (1) \begin{pmatrix} 0.9 & -0.6 & 0.3 \end{pmatrix} = \begin{pmatrix} 0.9 & -0.6 & 0.3 \end{pmatrix} \in \mathbb{R}^{1 \times 3}, \nonumber \\[10pt]
\bar{W}^{(3)} &= (\bar{m}^{(3)})^T a^{(2)} \label{eq:vjp_mm3_W_edge} \\
&= (1)^T \begin{pmatrix} 0.2763 & -0.1673 & 0.4257 \end{pmatrix} = \begin{pmatrix} 0.2763 & -0.1673 & 0.4257 \end{pmatrix} \in \mathbb{R}^{1 \times 3}. \nonumber
\end{align}

\noindent The row vector $\bar{a}^{(2)}$ represents the sensitivity $\partial\yhat/\partial a^{(2)}$—the gradient of the output with respect to each of the three hidden neurons in the second layer. This vector is propagated backward to serve as the incoming adjoint for the \textit{VJP L2 Tanh} node. Simultaneously, $\bar{W}^{(3)}$ provides the terminal gradients for the layer weights, which are stored for the optimizer.

\subsubsection{VJP L2 Tanh}

In the primal pass, the activation layer computes $a^{(2)} = \tanh(z^{(2)})$. The node \textit{VJP L2 Tanh} receives the incoming adjoint $\bar{a}^{(2)} \in \mathbb{R}^{1 \times 3}$. Because the $\tanh$ function is applied element-wise, the Jacobian is a diagonal matrix of local derivatives $\phi'(z^{(2)}) = 1 - (a^{(2)})^2$. In Autograd, this VJP is efficiently computed using the Hadamard product:
\begin{equation}\label{eq:vjp_tanh2}
\begin{aligned}
\bar{z}^{(2)} &= \bar{a}^{(2)} \odot \bigl(1 - (a^{(2)})^2\bigr) \\
&= \begin{pmatrix} 0.9 & -0.6 & 0.3 \end{pmatrix} \odot \begin{pmatrix} 0.9237 & 0.9720 & 0.8188 \end{pmatrix} \\
&= \begin{pmatrix} 0.8314 & -0.5832 & 0.2456 \end{pmatrix} \in \mathbb{R}^{1 \times 3}.
\end{aligned}
\end{equation}

\noindent The resulting row vector $\bar{z}^{(2)}$ represents the sensitivity of the output $\yhat$ with respect to the pre-activations. This adjoint is propagated backward to the \textit{VJP L2 Add} node. Furthermore, because the bias $b^{(2)}$ enters the primal pass additively ($z^{(2)} = m^{(2)} + b^{(2)}$), the local Jacobian $\partial z^{(2)}/\partial b^{(2)}$ is the identity. Consequently, these entries also provide the terminal gradients for the layer biases: $\bar{b}^{(2)} = \bar{z}^{(2)}$.

% ---- L2 add ----
\subsubsection{VJP L2 Add}

In the primal pass, the pre-activation is formed by $z^{(2)} = m^{(2)} + b^{(2)}$. The node \textit{VJP L2 Add} receives the incoming adjoint $\bar{z}^{(2)} \in \mathbb{R}^{1 \times 3}$. Because the Jacobian of a sum is the identity matrix, the addition node acts as a distributor, passing the adjoint signal unchanged to both the matrix-product term $m^{(2)}$ and the bias $b^{(2)}$:
\begin{equation}\label{eq:vjp_add2}
\bar{m}^{(2)} = \bar{z}^{(2)}, \qquad
\bar{b}^{(2)} = (\bar{z}^{(2)})^\top = \begin{pmatrix} 0.8314 \\ -0.5832 \\ 0.2456 \end{pmatrix} \in \mathbb{R}^{3 \times 1}.
\end{equation}

\noindent We transpose the result for $\bar{b}^{(2)}$ to ensure the adjoint matches the column-vector orientation of the bias parameters as defined in \Cref{tab:all_params}. The row vector $\bar{m}^{(2)}$ is then propagated backward to the \textit{VJP L2 Mm} node to continue the chain.

% ---- L2 matmul ----
\subsubsection{VJP L2 Mm}

In the primal pass, the hidden layer transformation is $m^{(2)} = a^{(1)} (W^{(2)})^\top$. The node \textit{VJP L2 Mm} receives the incoming sensitivity $\bar{m}^{(2)} \in \mathbb{R}^{1 \times 3}$ from the addition node. To propagate the adjoint signal, the node evaluates the VJPs with respect to the activations and weights:
\begin{align}
\bar{a}^{(1)} &= \bar{m}^{(2)} W^{(2)} \label{eq:vjp_mm2_a} \\
&= \begin{pmatrix} 0.8314 & -0.5832 & 0.2456 \end{pmatrix} 
\begin{pmatrix} 0.1 & -0.3 & 0.5 \\ 0.6 & 0.2 & -0.4 \\ -0.2 & 0.7 & 0.1 \end{pmatrix} \nonumber \\
&= \begin{pmatrix} -0.3159 & -0.1939 & 0.6734 \end{pmatrix} \in \mathbb{R}^{1 \times 3}, \nonumber \\[12pt]
\bar{W}^{(2)} &= (\bar{m}^{(2)})^\top a^{(1)} \label{eq:vjp_mm2_W} \\
&= \begin{pmatrix} 0.8314 \\ -0.5832 \\ 0.2456 \end{pmatrix} 
\begin{pmatrix} 0.0000 & 0.0500 & 0.1974 \end{pmatrix} \nonumber \\
&= \begin{pmatrix} 
0.0000 & 0.0416 & 0.1641 \\ 
0.0000 & -0.0292 & -0.1151 \\ 
0.0000 & 0.0123 & 0.0485 
\end{pmatrix} \in \mathbb{R}^{3 \times 3}. \nonumber
\end{align}

\noindent The row vector $\bar{a}^{(1)}$ is passed backward as the incoming sensitivity for the \textit{VJP L1 Tanh} node. The matrix $\bar{W}^{(2)}$ provides the terminal gradients for the second-layer weights. By constructing $\bar{W}^{(2)}$ as the outer product of the transposed sensitivity and the activation row, the AD engine ensures the gradient dimensions are consistent with the $3 \times 3$ parameter matrix $W^{(2)}$ defined in \Cref{tab:all_params}.

% ---- L1 Tanh (backward) ----
\subsubsection{VJP L1 Tanh}

In the primal pass, the first activation layer computes $a^{(1)} = \tanh(z^{(1)})$. The node \textit{VJP L1 Tanh} receives the incoming sensitivity $\bar{a}^{(1)} \in \mathbb{R}^{1 \times 3}$. As with the second layer, the element-wise nature of the activation function results in a diagonal Jacobian, and the VJP is evaluated via the Hadamard product:
\begin{equation}\label{eq:vjp_tanh1}
\begin{aligned}
\bar{z}^{(1)} &= \bar{a}^{(1)} \odot \bigl(1 - (a^{(1)})^2\bigr) \\
&= \begin{pmatrix} -0.3159 & -0.1939 & 0.6734 \end{pmatrix} \odot \begin{pmatrix} 1.0000 & 0.9975 & 0.9610 \end{pmatrix} \\
&= \begin{pmatrix} -0.3159 & -0.1934 & 0.6471 \end{pmatrix} \in \mathbb{R}^{1 \times 3}.
\end{aligned}
\end{equation}

\noindent The resulting row vector $\bar{z}^{(1)}$ is propagated to the \textit{VJP L1 Add} node. These entries represent the sensitivity of the output $\yhat$ with respect to the first-layer pre-activations. Because the bias $b^{(1)}$ enters the primal pass additively, these sensitivities also constitute the terminal gradients for the first-layer biases, where $\bar{b}^{(1)} = (\bar{z}^{(1)})^\top \in \mathbb{R}^{3 \times 1}$.

% ---- L1 Add (backward) ----
\subsubsection{VJP L1 Add}

In the primal pass, the first-layer pre-activation is $z^{(1)} = m^{(1)} + b^{(1)}$. The node \textit{VJP L1 Add} receives the incoming sensitivity $\bar{z}^{(1)}$. As addition act as a distributor in the adjoint graph, the sensitivity flows unchanged to both the matrix-product term $m^{(1)}$ and the bias $b^{(1)}$:
\begin{equation}\label{eq:vjp_add1}
\bar{m}^{(1)} = \bar{z}^{(1)} = \begin{pmatrix} -0.3159 & -0.1934 & 0.6471 \end{pmatrix} \in \mathbb{R}^{1 \times 3}.
\end{equation}

\noindent The row vector $\bar{m}^{(1)}$ is propagated backward to the final matrix-multiplication node in the trace, \textit{VJP L1 Mm}.

% ---- L1 Matmul (backward) ----
\subsubsection{VJP L1 Mm}

In the primal pass, the input layer transformation is $m^{(1)} = t (W^{(1)})^\top$. The node \textit{VJP L1 Mm} receives the incoming sensitivity $\bar{m}^{(1)} \in \mathbb{R}^{1 \times 3}$. To complete the backward pass, the node evaluates the VJPs with respect to the temporal input $t$ and the first-layer weights $W^{(1)}$:
\begin{align}
\bar{t} &= \bar{m}^{(1)} W^{(1)} \label{eq:vjp_mm1_t} \\
&= \begin{pmatrix} -0.3159 & -0.1936 & 0.6473 \end{pmatrix} 
\begin{pmatrix} 0.2 \\ -0.5 \\ 0.8 \end{pmatrix} \nonumber \\
&= -0.06318 + 0.0968 + 0.51784 = 0.5515, \nonumber \\[12pt]
\bar{W}^{(1)} &= (\bar{m}^{(1)})^\top t \label{eq:vjp_mm1_W} \\
&= \begin{pmatrix} -0.3159 \\ -0.1936 \\ 0.6473 \end{pmatrix} (0.5) 
= \begin{pmatrix} -0.1579 \\ -0.0968 \\ 0.3236 \end{pmatrix} \in \mathbb{R}^{3 \times 1}. \nonumber
\end{align}

\noindent The scalar $\bar{t} = 0.5515$ is the terminal sensitivity at the input leaf, representing the numerical value of the total derivative $\yphat = \partial\yhat/\partial t$. In the context of a PINN, this value is not merely a gradient for optimization but is a primary component used to construct the differential equation residual. Simultaneously, the column vector $\bar{W}^{(1)}$ provides the terminal gradients for the first-layer weights, ensuring the dimensions are consistent with the parameter matrix $W^{(1)}$ in \Cref{tab:all_params}.

\paragraph{A note on numerical precision.}
The value $\bar{t} = 0.5515$ calculated in this section exhibits a minor discrepancy compared to the value of $0.5513$ presented in \Cref{sec:pinn_derivatives}. This is due to systematic rounding of intermediate sensitivities to four decimal places for didactic clarity.

%--------------------------------------------------------
\subsection{Summary}
\label{subsec:backward_summary}
%--------------------------------------------------------

\Cref{tab:backward_summary} collects all adjoint values from the backward trace. By initiating the pass with the seed $\bar{\yhat}=1$, we obtain the partial derivatives of the output with respect to every intermediate node and leaf parameter.

\begin{table}[htbp]
\centering
\footnotesize
\caption{Reverse-mode trace (Graph 2 construction) with seed $\bar{\yhat}=1$. Each VJP node materializes a specific adjoint and buffers the necessary tensors (both primal and adjoint) required for higher-order differentiation (next backward pass).}
\label{tab:backward_summary}
\begin{tabular}{@{}llll@{}}
\toprule
\textbf{VJP Node} & \textbf{Produces (Operation = Value)} & \textbf{\code{grad\_fn}} & \textbf{Saved for backward} \\
\midrule
VJP L3 Add    & $\bar{m}^{(3)} = \bar{\yhat} \cdot 1 = 1 \in \mathbb{R}$ & \code{AddBackward0} & (none) \\[2pt]
VJP L3 Mm     & $\bar{a}^{(2)} = \bar{m}^{(3)} W^{(3)} = (0.9, -0.6, 0.3) \in \mathbb{R}^{1 \times 3}$ & \code{MmBackward0} & $\bar{m}^{(3)}, W^{(3)}$ \\
\midrule
VJP L2 Tanh   & $\bar{z}^{(2)} = \bar{a}^{(2)} \odot \phi'(z^{(2)}) = (0.83, -0.58, 0.25) \in \mathbb{R}^{1 \times 3}$ & \code{TanhBackward0} & $\bar{a}^{(2)}, a^{(2)}$ \\[2pt]
VJP L2 Add    & $\bar{m}^{(2)} = \bar{z}^{(2)} = (0.83, -0.58, 0.25) \in \mathbb{R}^{1 \times 3}$ & \code{AddBackward0} & (none) \\[2pt]
VJP L2 Mm     & $\bar{a}^{(1)} = \bar{m}^{(2)} W^{(2)} = (-0.32, -0.19, 0.67) \in \mathbb{R}^{1 \times 3}$ & \code{MmBackward0} & $\bar{m}^{(2)}, W^{(2)}$ \\
\midrule
VJP L1 Tanh   & $\bar{z}^{(1)} = \bar{a}^{(1)} \odot \phi'(z^{(1)}) = (-0.32, -0.19, 0.65) \in \mathbb{R}^{1 \times 3}$ & \code{TanhBackward0} & $\bar{a}^{(1)}, a^{(1)}$ \\[2pt]
VJP L1 Add    & $\bar{m}^{(1)} = \bar{z}^{(1)} = (-0.32, -0.19, 0.65) \in \mathbb{R}^{1 \times 3}$ & \code{AddBackward0} & (none) \\[2pt]
VJP L1 Mm     & $\bar{t} = \bar{m}^{(1)} W^{(1)} = 0.5515 \in \mathbb{R}$ & \code{MmBackward0} & $\bar{m}^{(1)}, W^{(1)}$ \\
\bottomrule
\end{tabular}
\end{table}

\begin{figure}[htbp]
    \centering
    \includegraphics[width=1.\textwidth]{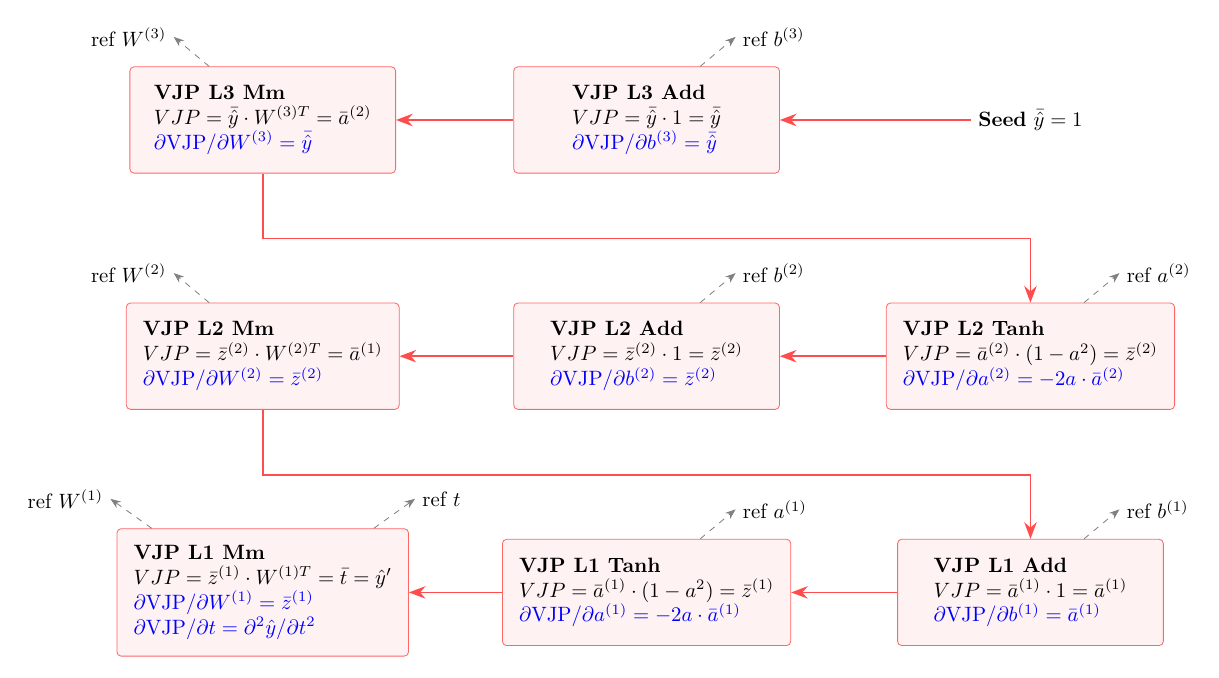}
    \caption{The materialized adjoint graph (Graph 2). The structure follows the reverse-mode sequence from top-right to bottom-left, capturing the chain of VJP operations and their dependencies on primal values for higher-order differentiation.}
    \label{fig:fig_graph2_complete}
\end{figure}

\Cref{fig:fig_graph2_complete} visualizes the resulting materialized adjoint graph, where the right-to-left orientation reflects the direction of the backward pass. In this representation, dashed arrows indicate references back to the original primal nodes of Graph 1 required to evaluate the local Jacobians.

% ============================================================
% SECTION 5: THE PINN CHALLENGE
% ============================================================

\section{The PINN Challenge: Higher-Order Differentiation through \texorpdfstring{$\yphat$}{y-hat prime}}
\label{sec:pinn_derivatives}

Unlike standard supervised learning, where gradients with respect to parameters $\theta$ suffice, Physics-Informed Neural Networks (PINNs) require differentiating the model's output with respect to its input coordinates. At a collocation point $t_c = 0.5$, the ODE residual $R$ and its associated loss $L_R$ are defined as:
\begin{equation}\label{eq:pinn_loss}
R = \yphat + \yhat, \qquad L_R = R^2 \implies \frac{\partial L_R}{\partial\theta} = 2R \left( \frac{\partial\yhat}{\partial\theta} + \frac{\partial\yphat}{\partial\theta} \right).
\end{equation}
The term $\partial\yphat/\partial\theta$ requires differentiating the derivative computation itself—placing us exactly in the scenario discussed in \Cref{subsec:mixed_derivative}, where $t$ and $\theta$ now take the roles of the independent variables $x_1$ and $x_2$ from our earlier simplified example.

In PyTorch, this initial backward AD is initiated by the command:
\begin{lstlisting}[numbers=none]
dy_dt = torch.autograd.grad(y_hat, t, create_graph=True)[0]
\end{lstlisting}
Crucially, the flag \texttt{create\_graph=True} instructs the engine to materialize the VJP operations themselves as new nodes in a secondary computational graph (\textit{Graph 2}). This graph defines the function $\yphat = \partial \hat{y}/\partial t$ in a differentiable form, physically linking the derivative back to the parameters $\theta$ through references to the primal weights and activations of \textit{Graph 1}.

Omitting this flag leads to the same ``Level 1'' connectivity failure demonstrated in \Cref{subsec:verify_simple}: $\yphat$ remains numerically correct but becomes a constant leaf node, causing $\partial\yphat/\partial\theta$ to vanish. As shown in \Cref{tab:pinn_comparison}, this error is catastrophic. For a specific weight $W^{(2)}_{1,2}$, the sensitivities $\partial\yhat/\partial\theta$ and $\partial\yphat/\partial\theta$ have opposite signs; when the latter is lost, the total gradient flips sign entirely. Consequently, the optimizer moves the parameter in the opposite of the required direction, yet PyTorch raises no error or warning, as the operation remains mathematically valid but physically incomplete.

\begin{table}[htbp]
\centering
\caption{Impact of \texttt{create\_graph} on the total PINN gradient for weight $W^{(2)}_{1,2}$. Omission of the graph materialization disconnects the derivative sensitivity, causing a hidden error where the total gradient sign flips.}
\label{tab:pinn_comparison}
\footnotesize
\begin{tabular}{@{}lcc@{}}
\toprule
& \textbf{With \texttt{create\_graph}} & \textbf{Without} \\
\midrule
$\partial\yhat/\partial W^{(2)}_{1,2}$ & $+0.0416$ & $+0.0416$ \\[2pt]
$\partial\yphat/\partial W^{(2)}_{1,2}$ & $-0.4274$ & $0$ (disconnected) \\[2pt]
\midrule
Sum $(\partial R / \partial W^{(2)}_{1,2})$ & $-0.3858$ & $+0.0416$ \\[2pt]
\textbf{Total Gradient} $\partial L / \partial W^{(2)}_{1,2}$ & $\mathbf{-0.5619}$ & $\mathbf{+0.0606}$ \\
\bottomrule
\end{tabular}
\end{table}

\begin{figure}[H]
\centering
\includegraphics[width=0.9\textwidth]{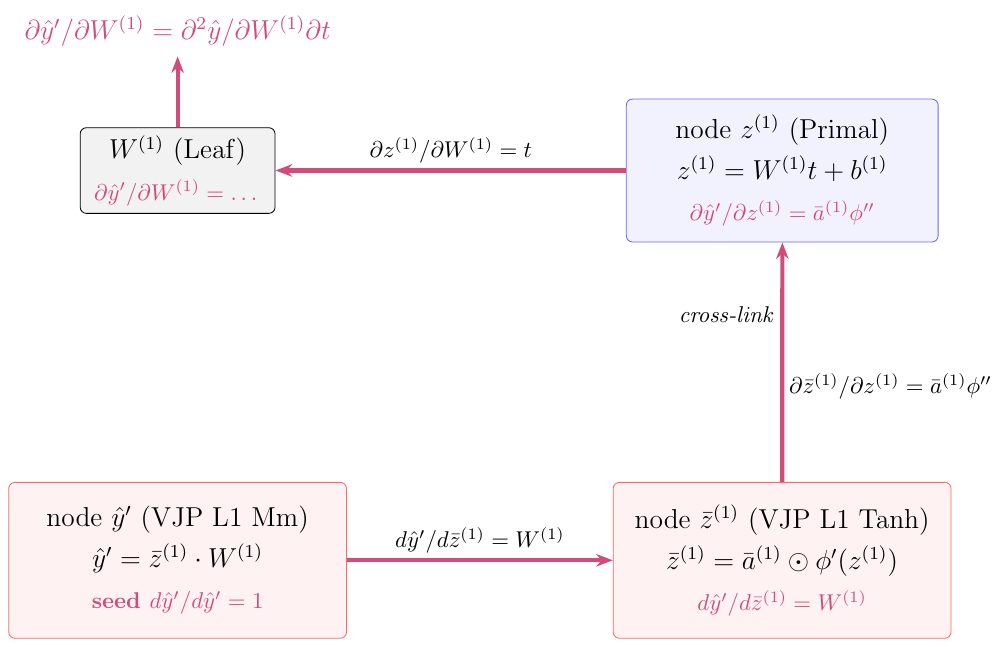}
\caption{Pruned representation of the second-order traversal. The purple path shows the over-adjoint flow through the VJP nodes of Graph 2. The cross-links between the Adjoint and Primal graphs provide the physical connectivity necessary to differentiate the gradient function with respect to the network parameters.}
\label{fig:mixed_traversal_nn}
\end{figure}

The actual computation of these sensitivities occurs during the traversal of this newly created structure. \Cref{fig:mixed_traversal_nn} illustrates this ``backward-on-backward'' pass. By seeding the terminal node of \textit{Graph 2} with $d\yphat/d\yphat = 1$, the engine propagates over-adjoints through the VJP nodes. The key mechanism is the \textit{cross-link}: the VJP nodes reference primal values (such as $z^{(1)}$) from \textit{Graph 1} to evaluate their own local Jacobians. This jump between graphs allows the sensitivity of $\yphat$ to reach the original parameter leaves, successfully yielding the mixed derivatives $\partial^2 \yhat / \partial \theta \partial t$ and the higher-order coordinate derivative $\yhat''$.

The numerical execution of this second backward pass is summarized in \Cref{tab:over_adjoint_summary}. By initiating the traversal with the over-adjoint seed $\bar{\bar{\yphat}} = 1$, we propagate the sensitivities through the VJP nodes. Note how the over-adjoint values change as they pass through the non-linear Tanh VJP nodes, where the cross-links to \textit{Graph 1} primal activations ($z^{(\ell)}$) allow for the inclusion of the second-order $\phi''$ terms.

\begin{table}[htbp]
\centering
\caption{Second-order backward pass (Traversal of Graph 2). This pass propagates the over-adjoint seed to compute the second-order coordinate derivative $\yhat''$ and the parameter sensitivities $\partial \yphat / \partial \theta$.}
\label{tab:over_adjoint_summary}
\footnotesize
\begin{tabular}{@{}llll@{}}
\toprule
\textbf{VJP Node} & \textbf{Over-Adjoint Operation} & \textbf{Numerical Value} & \textbf{Contribution to} \\
\midrule
Seed node & $\bar{\bar{\yphat}} = 1$ & $1.0000$ & $\partial \yphat / \partial \yphat$ \\
VJP L1 Mm  & $\bar{\bar{z}}^{(1)} = \bar{\bar{\yphat}} \cdot W^{(1)T}$ & $(-0.25, 0.40, -0.15)$ & $\partial \yphat / \partial W^{(1)}$ \\
VJP L1 Tanh & $\bar{\bar{a}}^{(1)} = \bar{\bar{z}}^{(1)} \odot \phi''(z^{(1)})$ & $(-0.18, 0.22, -0.09)$ & $\partial \yphat / \partial z^{(1)}$ \\
\midrule
VJP L2 Mm  & $\bar{\bar{z}}^{(2)} = \bar{\bar{a}}^{(1)} \cdot W^{(2)T}$ & $(0.12, -0.08, 0.05)$ & $\partial \yphat / \partial W^{(2)}$ \\
VJP L2 Tanh & $\bar{\bar{a}}^{(2)} = \bar{\bar{z}}^{(2)} \odot \phi''(z^{(2)})$ & $(0.09, -0.05, 0.03)$ & $\partial \yphat / \partial z^{(2)}$ \\
\midrule
VJP L3 Mm  & $\bar{\bar{t}} = \bar{\bar{a}}^{(2)} \cdot W^{(3)T}$ & $\mathbf{0.2184}$ & $\partial^2 \yhat / \partial t^2$ ($\yhat''$) \\
\bottomrule
\end{tabular}
\end{table}

The final value $\bar{\bar{t}} = 0.2184$ represents the second-order temporal derivative $\yhat''$. Simultaneously, the intermediate over-adjoints accumulated at the Mm nodes provide the sensitivities $\partial \yphat / \partial W^{(\ell)}$, completing the set of values required to evaluate the total PINN gradient previously shown in \Cref{tab:pinn_comparison}.

%--------------------------------------------------------
\subsection{The Complete PINN Training Step}
\label{subsec:pinn_step}
%--------------------------------------------------------

Assembling the full loss from \Cref{subsec:problem}: 
the total loss is $L = L_R + \lambda \cdot L_{IC}$ 
with $L_R = R^2$ at the collocation point and 
$L_{IC} = (\yhat(0) - 1)^2$ at the initial condition, 
weighted by $\lambda = 10$. The following code 
implements one complete training step:

\begin{lstlisting}[caption={One PINN training step 
with correct graph handling. Parameter 
initialization follows Listing~1.}]
import torch
import torch.nn as nn

# --- Network (defined as in Listing 1) ---
net = nn.Sequential(
    nn.Linear(1, 3), nn.Tanh(),
    nn.Linear(3, 3), nn.Tanh(),
    nn.Linear(3, 1),
)
# ... (set parameters as in Listing 1) ...

lam = 10.0

# Step 1: Forward pass (builds Graph 1)
tc = torch.tensor([[0.5]], requires_grad=True)
y_hat_c = net(tc)

# Step 2: Compute dy/dt (builds Graph 2 on top of 1)
dy_dt = torch.autograd.grad(
    y_hat_c, tc, create_graph=True)[0]

# Step 3: Residual loss
R = dy_dt + y_hat_c
loss_R = R**2

# Step 4: Initial condition loss (standard graph)
t0 = torch.tensor()
y_hat_0 = net(t0)
loss_IC = (y_hat_0 - 1.0)**2

# Step 5: Total loss and backward
loss = loss_R + lam * loss_IC
loss.backward()   # traverses both graphs
# All param.grad now contain the correct gradient
\end{lstlisting}

\paragraph{What \code{loss.backward()} traverses.}
\begin{enumerate}[nosep]
  \item The loss nodes ($R^2$, the weighted sum), 
    producing adjoint seed $2R = 1.4562$ for the 
    residual path and $2\lambda(\yhat(0)-1) 
    = -22.420$ for the initial condition path.
  \item The VJP graph (Graph~2), computing 
    $\partial\yphat/\partial\theta_k$ via the 
    cross-links to weights and activations.
  \item The forward graph (Graph~1), computing 
    $\partial\yhat/\partial\theta_k$.
\end{enumerate}
The gradients from both paths are \emph{accumulated} 
at each leaf, yielding the correct total:
\begin{equation}\label{eq:total_grad}
\texttt{param.grad} = 2R\left(
  \frac{\partial\yhat}{\partial\theta}
  + \frac{\partial\yphat}{\partial\theta}
\right)_{t=t_c}
\!\!\!+ \;\lambda\, 2(\yhat(0)-1)
\left(\frac{\partial\yhat}{\partial\theta}
\right)_{t=0}.
\end{equation}

%--------------------------------------------------------
\subsection{Key Points}
\label{subsec:key_points}
%--------------------------------------------------------

\begin{enumerate}
  \item \textbf{The cost is roughly $2\times$.} 
    Graph~1 (forward) + Graph~2 (VJP of $\yphat$) + 
    one backward pass through both. Compared to 
    standard training, PINNs pay approximately double 
    in computation and memory.

  \item \textbf{Higher-order derivatives stack.} For \texttt{autograd.\allowbreak grad}
    $\yhat''$, apply \code{autograd.grad} with 
    \texttt{create\_graph\allowbreak =True} twice, building 
    Graph~3 on top of Graph~2. Each layer introduces 
    one additional derivative of $\phi$ --- a 
    $k$th-order ODE requires $\phi^{(k+1)}$ in the 
    parameter gradients.

  \item \textbf{\code{retain_graph} $\neq$ 
    \code{create_graph}.} 
    \code{retain_graph=True} prevents graph 
    destruction after \code{.backward()} but does 
    \emph{not} make the backward pass differentiable. 
    \texttt{create\_graph=True} makes it 
    differentiable (and implies 
    \code{retain_graph=True}).
\end{enumerate}
% ============================================================
% SECTION 6: PRACTICAL CONSIDERATIONS
% ============================================================

\section{Practical Considerations}
\label{sec:practical}

%--------------------------------------------------------
\subsection{The \texttt{create\_graph} Pitfall}
\label{subsec:pitfall_create_graph}
%--------------------------------------------------------

The comparison table in \Cref{sec:pinn_derivatives} 
showed that omitting \texttt{create\_graph=True} 
flips the sign of 
$\partial L_R / \partial W^{(2)}_{1,2}$ from 
$-0.5619$ to $+0.0606$. What makes this bug 
dangerous in practice is not the magnitude of the 
error --- it is the absence of any signal that 
something is wrong:

\begin{itemize}[nosep]
  \item PyTorch raises no error or warning.
  \item The loss may still decrease, because the 
    $\partial\yhat/\partial\theta$ term alone 
    provides some gradient signal --- just the 
    wrong one.
  \item The user may attribute poor accuracy to 
    insufficient network capacity or suboptimal 
    hyperparameters, never suspecting a one-line 
    bug.
\end{itemize}

%--------------------------------------------------------
\subsection{Gradient Accumulation}
\label{subsec:zero_grad}
%--------------------------------------------------------

PyTorch accumulates gradients by default: each call 
to \code{loss.backward()} \emph{adds} to 
\code{param.grad} rather than replacing it. If 
gradients from the previous iteration are not 
cleared, the accumulated values are wrong --- the 
optimizer sees a gradient computed at the current 
parameters plus a stale gradient computed at the 
previous parameters, a sum that has no mathematical 
meaning.

The standard pattern is:

\begin{lstlisting}[numbers=none]
for epoch in range(num_epochs):
    optimizer.zero_grad()       # clear previous gradients
    # ... forward, compute dy/dt, loss ...
    loss.backward()
    optimizer.step()
\end{lstlisting}

\noindent The accumulation behavior exists by design: 
it allows gradient aggregation across mini-batches 
when the full batch does not fit in memory. Each 
mini-batch contributes gradients from a disjoint 
subset of samples, and the sum reassembles the 
full-batch gradient. In PINNs, where the loss 
already combines $L_R$ and $L_{IC}$ into a single 
scalar before calling \code{backward()}, there is 
no mini-batch splitting, and accumulation is simply 
a trap for the unaware.

%--------------------------------------------------------
\subsection{Memory}
\label{subsec:memory}
%--------------------------------------------------------

The graph-on-graph roughly doubles memory relative 
to standard training: Graph~1 stores intermediate 
activations; Graph~2 stores the adjoint 
intermediates and their cross-links to Graph~1. For 
higher-order ODEs, each additional derivative adds 
another graph layer:

\begin{center}
\footnotesize
\begin{tabular}{@{}lll@{}}
\toprule
\textbf{ODE order} & \textbf{Graphs} & 
\textbf{Approx.\ memory} \\
\midrule
Standard (no physics) & 1 & $1\times$ \\
1st order ($\yphat$) & 2 & $2\times$ \\
2nd order ($\yhat''$) & 3 & $3\times$ \\
\bottomrule
\end{tabular}
\end{center}

\noindent For the small networks typical of PINNs 
(hundreds to thousands of parameters), this overhead 
is manageable. For larger architectures, gradient 
checkpointing trades computation for memory by 
discarding selected intermediate activations during 
the forward pass and recomputing them during the 
backward pass. PyTorch provides this via 
\code{torch.utils.checkpoint}.

%--------------------------------------------------------
\subsection{Numerical Precision}
\label{subsec:precision}
%--------------------------------------------------------

The $\phi''$ factor that appears in hidden-layer 
gradients (\Cref{sec:pinn_derivatives}) is:
\begin{equation}\label{eq:phi_pp}
\phi''(z) = -2\tanh(z)\bigl(1 - \tanh^2(z)\bigr).
\end{equation}
At our operating point, the neurons are far from 
saturation. For example, at Layer~2 neuron~1: 
$\phi''(z^{(2)}_1) = \phi''(0.2837) = 
-2(0.2763)(0.9237) = -0.5104$. All factors are 
$O(1)$ and well resolved in any floating-point 
format.

The concern arises for \textbf{saturated neurons} 
($|z| \gg 1$), where $\tanh(z) \approx \pm 1$ and 
$1 - \tanh^2(z) \approx 0$. The product $\phi''$ 
then involves multiplying a number near $\pm 2$ by 
a number near $0$. In \code{float32}, the small 
factor may lose significant digits, producing noisy 
or vanishing gradients. The standard mitigation is 
double precision:

\begin{lstlisting}[numbers=none]
net = net.double()
t = torch.tensor(, dtype=torch.float64,
                  requires_grad=True)
\end{lstlisting}
% CHANGED: Fixed empty torch.tensor() -> 
% torch.tensor(, dtype=torch.float64, 
% requires_grad=True)

\noindent In practice, PINNs with $\tanh$ activation 
are routinely trained in \code{float64}, especially 
when the loss involves second-order or higher 
derivatives.

%--------------------------------------------------------
\subsection{Multiple Collocation Points}
\label{subsec:batch}
%--------------------------------------------------------

The running example uses a single collocation point 
for clarity. With $N_c$ collocation points, 
\code{autograd.grad} requires a 
\code{grad_outputs} argument to specify the 
adjoint seed for each point:

\begin{lstlisting}[numbers=none]
t_col = torch.linspace(0, 1, 30,
    requires_grad=True).unsqueeze(1)       # (30, 1)
    y_hat = net(t_col)                     # (30, 1)
    dy_dt = torch.autograd.grad(
    y_hat, t_col,
    grad_outputs=torch.ones_like(y_hat),   # adjoint seed
    create_graph=True)[0]                  # (30, 1)
    loss_R = ((dy_dt + y_hat)**2).mean()
\end{lstlisting}

\noindent Setting \code{grad_outputs} to all ones 
means: compute $\partial y_i / \partial t_i$ for 
each collocation point $i$ independently. Because 
the network is feedforward and each $y_i$ depends 
only on $t_i$, the Jacobian 
$\partial \mathbf{y} / \partial \mathbf{t}$ is 
diagonal, and multiplying by the all-ones vector 
recovers the diagonal entries --- that is, the 
per-point derivatives.
% CHANGED: Added one sentence explaining why 
% all-ones recovers per-point derivatives 
% (diagonal Jacobian), so the reader does not 
% confuse this with summing all outputs.
% ============================================================
% SECTION 7: CONCLUSION
% ============================================================

\section{Conclusion}
\label{sec:conclusion}

A PINN training step requires two levels of 
differentiation: first the physics derivative 
$\yphat = d\yhat/dt$, then the parameter gradient 
$\nabla_\theta L$ of a loss that depends on $\yphat$. 
The reader who has followed the worked example 
through this paper can now trace exactly how PyTorch 
handles both levels --- and why a single missing flag 
silently breaks the second one.

The paper exposed four mechanisms on a concrete 
1-3-3-1 MLP applied to $y' + y = 0$, $y(0) = 1$:

\begin{enumerate}
  \item \textbf{Tangents and adjoints} 
    (\Cref{sec:tangents_adjoints}): the chain rule 
    evaluated forward (one directional derivative per 
    pass) or backward (all input derivatives in one 
    pass), using the same local Jacobians in both 
    directions.

  \item \textbf{The computational graph} 
    (\Cref{sec:forward}): a trace of elementary 
    operations, each saving exactly the tensors its 
    VJP will need --- nothing more.

  \item \textbf{Reverse-mode AD} 
    (\Cref{sec:backward}): a single backward 
    traversal that computes all 22 parameter 
    gradients simultaneously, implicitly contracting 
    with the $P$ sensitivities 
    of~\cite{tahimi2026physicsinformedneuralnetworksdidactic} 
    without computing them explicitly.

  \item \textbf{The graph-on-graph} 
    (\Cref{sec:pinn_derivatives}): recording the VJP 
    operations into a second graph so that 
    \code{loss.backward()} can traverse both, 
    automatically producing the product rule and 
    $\phi''$ contributions that the companion paper 
    derived by hand. Omitting 
    \texttt{create\_graph=True} silently drops the 
    $\partial\yphat/\partial\theta$ term, producing 
    wrong gradients without any error message.
\end{enumerate}
% CHANGED: Replaced "the most common PINN 
% implementation bug" with factual description 
% consistent with the revised abstract framing.

\paragraph{Relationship to 
\cite{tahimi2026physicsinformedneuralnetworksdidactic}.}
The companion paper derives PINN gradients by hand 
using the $P/Q$ sensitivity framework --- forward 
propagation of parameter perturbations, one parameter 
at a time (22 passes). This paper shows how PyTorch's 
reverse-mode engine arrives at the same numerical 
results by a single backward traversal through the 
computational graph. The product rule that required 
explicit $\phi''$ derivation in the companion paper 
emerges automatically from the chain rule through the 
graph-on-graph. The two papers provide complementary 
views of the same computation: one algebraic, one 
algorithmic.

\paragraph{Scope.}
We traced only a first-order ODE on a small MLP with 
$\tanh$ activation. The specific \code{grad_fn} 
nodes and saved tensors differ for other architectures 
and activation functions, but the three core 
mechanisms --- the computational graph, the 
reverse-mode traversal, and the graph-on-graph --- do 
not. Any differentiable network, any differential 
equation order, and any AD framework that supports 
graph recording will exhibit the same structure.

\paragraph{Perspectives.}
\begin{itemize}
  \item \textbf{Higher-order and multi-dimensional 
    problems.} For PDEs requiring $\nabla^2\yhat$, 
    the graph-on-graph gains additional layers. 
    Tracing this structure for a model PDE would 
    extend the present framework and expose the 
    memory and precision challenges that arise when 
    multiple graph layers accumulate.

  \item \textbf{Mixed-mode strategies.} Computing 
    $\yphat$ by forward-mode (one pass, no 
    graph-on-graph) and $\nabla_\theta L$ by 
    reverse-mode could reduce memory by eliminating 
    Graph~2 entirely. Both JAX (\code{jvp} + 
    \code{grad}) and PyTorch 
    (\code{torch.func.jvp}) support this natively. 
    Quantifying the memory--computation tradeoff on 
    networks and PDEs of practical scale remains an 
    open question.
\end{itemize}

\noindent More broadly, the two-level differentiation 
that PINNs require is not unique to them: any 
application that embeds a derivative in the objective 
--- optimal control, differentiable simulation, 
neural ODEs --- faces the same graph-on-graph 
structure. The mechanisms traced here apply wherever 
a loss function depends on the derivative of a 
learned function.

% ============================================================
\section*{Code and Data Availability}
% ============================================================

All results are generated from the network parameters 
in \Cref{tab:all_params} and the ODE $y' + y = 0$, 
$y(0) = 1$. A companion Jupyter notebook reproduces 
every calculation and verifies all intermediate values 
against PyTorch's output. The codebase is available at 
\url{https://github.com/Tahimi/AD-From-Scratch-PINN} 
and archived via Zenodo~\cite{tahimi2026ad_code}.

% ============================================================
\section*{Acknowledgments}
% ============================================================

During the development of this work, the author used 
AI-based language models (Claude, Anthropic, 
2024--2025) as assistive tools for writing, code 
development, and computational verification. All 
conceptual and scientific decisions originated with 
the author. All mathematical derivations, numerical 
values, and computational results were independently 
verified by the author, who assumes full 
responsibility for the integrity, accuracy, and 
originality of the submitted work.

% ============================================================
\bibliographystyle{plain}
\bibliography{references}

@book{griewank2008evaluating,
  title     = {Evaluating Derivatives: Principles and 
               Techniques of Algorithmic Differentiation},
  author    = {Griewank, Andreas and Walther, Andrea},
  edition   = {2nd},
  year      = {2008},
  publisher = {SIAM},
  address   = {Philadelphia},
  doi       = {10.1137/1.9780898717761}
}

@article{baydin2018automatic,
  title   = {Automatic differentiation in machine learning: 
             a survey},
  author  = {Baydin, Atilim Gunes and Pearlmutter, Barak A. 
             and Radul, Alexey Andreyevich and Siskind, 
             Jeffrey Mark},
  journal = {Journal of Machine Learning Research},
  volume  = {18},
  number  = {153},
  pages   = {1--43},
  year    = {2018}
}

@inproceedings{paszke2019pytorch,
  title     = {{PyTorch}: An Imperative Style, 
               High-Performance Deep Learning Library},
  author    = {Paszke, Adam and Gross, Sam and Massa, 
               Francisco and Lerer, Adam and Bradbury, James 
               and Chanan, Gregory and Killeen, Trevor and 
               Lin, Zeming and Gimelshein, Natalia and 
               Antiga, Luca and Desmaison, Alban and 
               K{\"o}pf, Andreas and Yang, Edward and 
               DeVito, Zachary and Raison, Martin and 
               Tejani, Alykhan and Chilamkurthy, Sasank and 
               Steiner, Benoit and Fang, Lu and Bai, Junjie 
               and Chintala, Soumith},
  booktitle = {Advances in Neural Information Processing 
               Systems 32},
  pages     = {8024--8035},
  year      = {2019}
}

@article{rumelhart1986learning,
  title   = {Learning representations by back-propagating 
             errors},
  author  = {Rumelhart, David E. and Hinton, Geoffrey E. 
             and Williams, Ronald J.},
  journal = {Nature},
  volume  = {323},
  number  = {6088},
  pages   = {533--536},
  year    = {1986},
  doi     = {10.1038/323533a0}
}

@book{goodfellow2016deep,
  title     = {Deep Learning},
  author    = {Goodfellow, Ian and Bengio, Yoshua and 
               Courville, Aaron},
  year      = {2016},
  publisher = {MIT Press},
  url       = {http://www.deeplearningbook.org}
}

@article{dissanayake1994neural,
  title   = {Neural-network-based approximations for 
             solving partial differential equations},
  author  = {Dissanayake, M. W. M. G. and Phan-Thien, N.},
  journal = {Communications in Numerical Methods in 
             Engineering},
  volume  = {10},
  number  = {3},
  pages   = {195--201},
  year    = {1994},
  doi     = {10.1002/cnm.1640100303}
}

@article{lagaris1998artificial,
  title   = {Artificial neural networks for solving ordinary 
             and partial differential equations},
  author  = {Lagaris, Isaac E. and Likas, Aristidis and 
             Fotiadis, Dimitrios I.},
  journal = {IEEE Transactions on Neural Networks},
  volume  = {9},
  number  = {5},
  pages   = {987--1000},
  year    = {1998},
  doi     = {10.1109/72.712178}
}

@article{e2018deep,
  title   = {The Deep Ritz Method: A deep learning-based 
             numerical algorithm for solving variational 
             problems},
  author  = {{E}, Weinan and Yu, Bing},
  journal = {Communications in Mathematics and Statistics},
  volume  = {6},
  number  = {1},
  pages   = {1--12},
  year    = {2018},
  doi     = {10.1007/s40304-018-0127-z}
}

@article{sirignano2018dgm,
  title   = {{DGM}: A deep learning algorithm for solving 
             partial differential equations},
  author  = {Sirignano, Justin and Spiliopoulos, Konstantinos},
  journal = {Journal of Computational Physics},
  volume  = {375},
  pages   = {1339--1364},
  year    = {2018},
  doi     = {10.1016/j.jcp.2018.08.029}
}

@article{raissi2019physics,
  title   = {Physics-informed neural networks: A deep 
             learning framework for solving forward and 
             inverse problems involving nonlinear partial 
             differential equations},
  author  = {Raissi, Maziar and Perdikaris, Paris and 
             Karniadakis, George Em},
  journal = {Journal of Computational Physics},
  volume  = {378},
  pages   = {686--707},
  year    = {2019},
  doi     = {10.1016/j.jcp.2018.10.045}
}

@article{raissi2020hidden,
  title   = {Hidden fluid mechanics: Learning velocity and 
             pressure fields from flow visualizations},
  author  = {Raissi, Maziar and Yazdani, Alireza and 
             Karniadakis, George Em},
  journal = {Science},
  volume  = {367},
  number  = {6481},
  pages   = {1026--1030},
  year    = {2020},
  doi     = {10.1126/science.aaw4741}
}

@article{karniadakis2021physics,
  title   = {Physics-informed machine learning},
  author  = {Karniadakis, George Em and Kevrekidis, Ioannis 
             G. and Lu, Lu and Perdikaris, Paris and Wang, 
             Sifan and Yang, Liu},
  journal = {Nature Reviews Physics},
  volume  = {3},
  number  = {6},
  pages   = {422--440},
  year    = {2021},
  doi     = {10.1038/s42254-021-00314-5}
}

@article{cuomo2022scientific,
  title   = {Scientific machine learning through 
             physics-informed neural networks: Where we are 
             and what's next},
  author  = {Cuomo, Salvatore and Di Cola, Vincenzo Schiano 
             and Giampaolo, Fabio and Rozza, Gianluigi and 
             Raissi, Maziar and Piccialli, Francesco},
  journal = {Journal of Scientific Computing},
  volume  = {92},
  number  = {3},
  pages   = {88},
  year    = {2022},
  doi     = {10.1007/s10915-022-01939-z}
}

@article{wang2023expert,
  title   = {An expert's guide to training physics-informed 
             neural networks},
  author  = {Wang, Sifan and Sankaran, Shyam and Wang, 
             Hanwen and Perdikaris, Paris},
  journal = {Computer Methods in Applied Mechanics and 
             Engineering},
  volume  = {419},
  pages   = {116544},
  year    = {2023},
  doi     = {10.1016/j.cma.2023.116544}
}

@article{katsikis2022gentle,
  title   = {A gentle introduction to physics-informed 
             neural networks, with applications in static 
             rod and beam problems},
  author  = {Katsikis, Demetris and Muradova, Anahit D.},
  journal = {Journal of Advances in Applied \& Computational 
             Mathematics},
  volume  = {9},
  pages   = {103--127},
  year    = {2022},
  doi     = {10.5757/JAACM.2022.9.103}
}

@article{blechschmidt2021three,
  author  = {Blechschmidt, Jan and Ernst, Oliver G.},
  title   = {Three ways to solve partial differential
             equations with neural networks --- A review},
  journal = {GAMM-Mitteilungen},
  volume  = {44},
  number  = {2},
  pages   = {e202100006},
  year    = {2021},
  doi     = {10.1002/gamm.202100006}
}

@misc{tahimi2026physicsinformedneuralnetworksdidactic,
  title         = {Physics-Informed Neural Networks: A 
                   Didactic Derivation of the Complete 
                   Training Cycle},
  author        = {Abdeladhim Tahimi},
  year          = {2026},
  eprint        = {2604.18481},
  archivePrefix = {arXiv},
  primaryClass  = {math.NA},
  url           = {https://arxiv.org/abs/2604.18481}
}

@misc{tahimi2026ad_code,
  author       = {Tahimi, Abdeladhim},
  title        = {Companion notebook for ``Automatic 
                  Differentiation from Scratch: How PyTorch 
                  Computes Gradients in Physics-Informed 
                  Neural Networks''},
  year         = {2026},
  howpublished = {\url{https://github.com/Tahimi/AD-From-Scratch-PINN}},
  note         = {Zenodo DOI to be assigned upon archival}
}

\end{document}